\documentclass{article}

\usepackage[preprint,nonatbib]{neurips_2025}

\usepackage[nonatbib]{neurips_2025}

\usepackage[utf8]{inputenc} 
\usepackage[T1]{fontenc}    
\usepackage{hyperref}       
\usepackage{url}            
\usepackage{booktabs}       
\usepackage{amsfonts}       
\usepackage{todonotes}      
\usepackage{nicefrac}       
\usepackage{microtype}      
\usepackage{xcolor}         
\usepackage{wrapfig}
\usepackage{graphicx}
\usepackage{algorithm}
\usepackage{algpseudocode}
\usepackage{amsmath}
\usepackage{adjustbox}
\usepackage{float}
\usepackage{multicol}
\usepackage{multirow}
\usepackage{subcaption}
\usepackage{tablefootnote}
\usepackage{inconsolata}
\usepackage{enumitem}
\usepackage{hyperref}

\title{TG-NAS: Generalizable Zero-Cost Proxies with Operator Description Embedding and Graph Learning for Efficient Neural Architecture Search}

\author{
  Ye Qiao\textsuperscript{1}, Jingcheng Li\textsuperscript{2}, Haocheng Xu\textsuperscript{1}, and Sitao Huang\textsuperscript{1}\\
  \textsuperscript{1}Department of EECS, University of California, Irvine\\
  \textsuperscript{2}Department of Computer Science, University of California, San Diego\\
  \texttt{\{yeq6, haochx, sitaoh\}@uci.edu}\\\texttt{jil458@ucsd.edu }
}

\begin{document}

\maketitle

\begin{abstract}
Neural Architecture Search (NAS) is a powerful technique for discovering high-performing CNN architectures, but most existing methods rely on costly training or extensive sampling. Zero-shot NAS offers a training-free alternative by using proxies to predict architecture performance. However, existing proxies are often suboptimal—frequently outperformed by simple metrics like parameter count or FLOPs—and they generalize poorly across different search spaces. Moreover, current model-based proxies struggle to adapt to new operators without access to ground-truth accuracy, limiting their transferability. We propose TG-NAS, a universal, model-based zero-cost (ZC) proxy that combines a Transformer-based operator embedding generator with a Graph Convolutional Network (GCN) to predict architecture performance. Unlike prior model-based predictors, TG-NAS requires no retraining and generalizes across arbitrary search spaces. It serves as a standalone ZC proxy with strong data efficiency, robustness, and cross-space consistency. Extensive evaluations across diverse NAS benchmarks demonstrate TG-NAS’s superior rank correlation and generalizability compared to existing proxies. Additional, it improves search efficiency by up to 300× and discovers architectures achieving 93.75\% CIFAR-10 accuracy on NAS-Bench-201 and 74.9\% ImageNet top-1 accuracy on the DARTS space, establishing TG-NAS as a promising foundation for efficient, generalizable NAS.
\end{abstract}

\section{Introduction}
\label{sec:intro}

Deep convolutional neural networks (CNN) have achieved incredible performance in computer vision, speech recognition, object detection, and many other fields \cite{krizhevsky2012imagenet, lin2015microsoft, qiao2022two, kaeley2023support}. Since then many manually designed CNN topologies that target either high performance or high computational efficiency have been proposed \cite{he2016deep,sandler2019mobilenetv2}. With larger and deeper expert-designed CNN models, people can easily achieve state-of-the-art performance in various vision tasks. 
However, manual neural network architecture design is costly and requires a significant amount of time, expertise, and computing resources. 
It becomes even more challenging if the neural network is designed for hardware platforms with stringent resource and energy constraints. 
To address the challenges in neural architecture design,  neural architecture search (NAS) was proposed to automate the CNN model design flow and explore specialized model architectures given application-specific performance and computing resource constraints \cite{lin2020mcunet, zoph2017neuralarchitecturesearchreinforcement}. Despite the advancements in the automated search techniques, the majority of neural architecture search (NAS) approaches still face the challenge of time-consuming training and evaluation. In the early stage, prevailing NAS methods typically involved iteratively sampling DNN architectures from the search space, training them, and using their performance to guide the search process \cite{muNAS}. These NAS approaches were primarily based on reinforcement learning (RL) or evolutionary algorithms.
\begin{wrapfigure}{R}{0.3\textwidth}
    \centering
        \includegraphics[width=\linewidth]{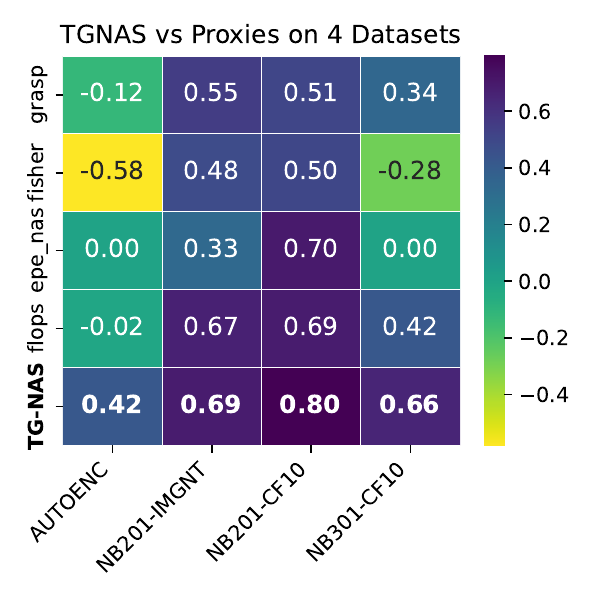}
    \caption{Spearman rank correlation comparison between ours and 4 other ZC proxies (the higher the better).}
    \label{fig:zcp_corr_4x4}
\end{wrapfigure}
In recent years, zero-shot proxies \cite{white2023neural, krishnakumar2022nasbenchsuitezero, mellor2021neural, qiao2024micronas, qiao2024monas} have emerged as a promising technique in neural architecture search (NAS). Zero-shot proxies use lightweight computation, typically a single forward pass of the DNN with one minibatch of data, to assign a score to each DNN architecture in the search space. The scores obtained from these proxies are expected to correlate with the final accuracy of candidate DNN architectures \cite{white2023neural, krishnakumar2022nasbenchsuitezero, mellor2021neural}. This approach not only conserves computational resources but also accelerates the overall NAS process, enabling researchers to explore a larger number of DNN architectures. However, recent general zero-shot proxies struggle to consistently perform well for practical uses. Many of them cannot outperform a na\"ive proxy like the number of parameters ($params$) or floating point operation count (\textit{FLOPs}) as shown in Figure \ref{fig:zcp_corr_4x4}, and they are often data-dependent and not fast enough in computation and also do not generalized well across various search space and downstream tasks. Furthermore, the generally high dependence and correlation of different zero-cost (ZC) proxies from similar domains, such as gradient information or kernel methods \cite{lee2019snip, dong2019searching}, also reduce the effectiveness of those zero-shot NAS approaches.

To address these limitations of zero-cost performance indicators, many model-based performance predictors have emerged. 
These predictors rely on training models to forecast the final validation accuracy of an architecture solely based on its topology and operators. 
Common models include Gaussian processes \cite{levesque2017bayesian}, deep neural networks \cite{shi2020bridging}, among others. However, these methods are not zero-shot and cannot be considered zero-cost since they typically require hundreds of fully-trained architectures as training data, resulting in prolonged search times. Additionally, model-based prediction has often been used only to accelerate NAS query times, given their lack of generalization and inability to handle unseen operators that are not encoded during the training of the predictor model.

This leads to fundamental research questions: how can we develop a pure universal prediction model capable of predicting the performance of any given DNN architecture? Is it possible to develop a generally applicable model-based predictor that can handle the unseen operators and mitigate generalization limitations aforementioned? 
In other words, can we make model-based predictors become a strong standalone zero-cost proxy? 
Our response to these questions is TG-NAS, a model-based predictor NAS proxy that can be generalized to new search spaces and previously unseen operators \textit{without} retraining the predictor model. It features a general Transformer-based operator encoder and a graph convolution network (GCN) predictor. TG-NAS provides an efficient and zero-cost (ZC) proxy solution to general neural architecture search problems and bridges the gap between model-based predictors and ZC proxies. We will open source this work to facilitate future research. The link to the source code can be found in the supplementary materials. 
The main contributions of TG-NAS can be summarized as follows:
\begin{itemize}[noitemsep, topsep=1pt]
  \item We propose \textbf{TG-NAS}, a universally applicable, data-independent performance predictor that generalizes to unseen operators and new search spaces without retraining. 
  TG-NAS introduces a novel architecture encoding strategy that integrates: (i) text-based operator descriptions processed by a fine-tuned sentence transformer to capture semantic relationships between operators, (ii) a directed acyclic graph (DAG) based representation of model architectures, and (iii) a GCN-based architecture performance predictor.
  \item TG-NAS serves as a zero-cost proxy, enabling prediction-only NAS with no training data dependency. We provide comprehensive analyses spanning 12 NAS benchmarks to demonstrate its superior performance, generalizability and proxy independence.
  \item Our analysis of TG-NAS proxy demonstrates remarkable performance on both NAS-Bench-201 and DARTS spaces with CIFAR-10 and ImageNet datasets. TG-NAS achieves up to $300\times$ faster search over SOTA zero-cost proxies and magnitudes of speedup over existing NAS methods, while maintaining high accuracy.

\end{itemize}

\section{Background and Related Work}
\subsection{Neural Architecture Search (NAS)}
NAS is an important technique in automating the design of neural architectures for a given task \cite{white2023neural}. A typical NAS contains a \textbf{search strategy} that selects the candidate architectures from a predefined \textbf{search space} and an \textbf{estimation strategy} that enables the performance estimation for candidates. 

\textbf{Search space} can be categorized as the following types: macro search space \cite{kandasamy2019neural}, chain-structures search space \cite{sandler2019mobilenetv2}, cell-based search space \cite{liu2018darts,ying2019nasbench101,dong2020nasbench201}, and hierarchical search space \cite{liu2018hierarchical}. Among those, cell-based search space is the most popular one in NAS \cite{white2023neural}. The searchable cells (a directed acyclic graph (DAG) of operations) make up the microstructure of the search space while the macrostructure 
(that defines the number of cells and how they stack together)
is fixed. For example, NAS-Bench-101 \cite{ying2019nasbench101} contains 423,624 unique architectures, and each cell consists of 7 nodes (each node is chosen from three operators). In NAS-Bench-201 \cite{dong2020nasbench201}, there are 15,625 cell candidates and each cell consists of four nodes (each node chosen from five operators).
In contrast, the DARTS search space \cite{liu2018darts} is more expansive, featuring approximately $10^{18}$ architectures. It consists of two cells, each containing seven nodes. The first two nodes receive inputs from previous layers, and the following four nodes can be any DAG structure, each having two incident edges. The last node serves as the output node, and each edge can take on one of eight operations.
In this work, we perform our experiments on those three search spaces to evaluate our proposed TG-NAS flow.  

The \textbf{search strategy} in NAS has been widely explored. There are well-known black-box optimizations, such as random selection with full training, reinforcement learning, evolution, Bayesian optimization, Monte Carlo tree search, etc. With these search strategies, we still need to train the searched architecture and use the performance result to guide the search, which is a time-consuming process. To overcome the training time bottleneck, one-shot techniques were introduced as an \textbf{estimation strategy}. 
These techniques involve
training a single (one-shot) supernet, which is an over-parameterized architecture that encompasses all possible architectures in the search space \cite{cai2019proxylessnas,liu2018darts}. Once the supernet is trained, each architecture in the search space can be evaluated by inheriting its weights from sampling the subnet within the supernet. Supernet design and training often become the performance bottleneck of these approaches. Supernet training typically dominates the NAS runtime. 

\subsection{Zero-Cost (ZC) Proxies for NAS}
To speed up the entire NAS process, a more efficient \textbf{estimation strategy} to predict the performance of the searched architecture is needed. Recently, zero-cost proxies have been introduced as a family of performance prediction techniques. 
The idea is to avoid training any neural network and just run a fast computation, e.g., one single forward pass of the candidate architecture over one tiny set of data, based on which assign a ranking or score to the candidate architecture. 



The expressivity of a neural network, which relates to the complexity of the function it can represent, has been widely used as one of the zero-cost proxies. Several recent works, such as TE-NAS \cite{chen2021neural}, Zen-NAS \cite{lin2021zennas}, and NASWOT \cite{mellor2021neural}, approximate the behavior of neural networks by considering the expressivity of ReLU network and the number of linear regions they can separate. Kernel methods have also been employed to approximate the convergence and generalization ability of networks without training. For instance, TE-NAS \cite{chen2021neural}  formulates neural networks as a Gaussian process and analyzes randomly-initialized architectures by the spectrum of the neural tangent kernel (NTK) \cite{jacot2020neural,xiao2020disentangling} and the number of linear regions in the input space. Zico \cite{li2023zico} extends such kernel-based analysis to reveal the relationships among the gradient properties, the training convergence, and generalization capacity of neural networks. Similarly, gradient with respect to the parameters of neural networks is proved to be different approximations of Taylor expansion of deep neural networks \cite{lee2019snip,wang2020picking,abdelfattah2021zerocost}. SNIP \cite{lee2019snip} introduced a saliency metric computed at initialization using a single minibatch of data; Grasp \cite{wang2020picking} improved upon SNIP \cite{lee2019snip} by approximating the change in gradient norm instead of loss; Synflow \cite{abdelfattah2021zerocost} generalized previous approaches and proposed a modified version computes a loss which is simply the product of all parameters in the network thus no data is needed. Gradient concerning feature maps combined with the number of linear regions have been used in Zen-NAS \cite{lin2021zennas}. There are also other zero-cost proxies leveraging Jacobian covariances, Fisher information, and other fundamental theoretical indicators \cite{turner2020blockswap, mellor2021neural,abdelfattah2021zerocost}. Although zero-cost proxies can expedite estimation time, they often compromise prediction accuracy and introduce biases stemming from data dependency \cite{white2023neural}.
\subsection{Challenges in Model-based Prediction}
Despite the inherent limitations of zero-cost performance prediction, the integration of model-based prediction has emerged as a pivotal component in guiding neural architecture search (NAS) algorithms. This approach is particularly useful when combined with Bayesian optimization and utilized as a subroutine \cite{kandasamy2019neural, shi2020bridging}. Various types of predictor models, including Gaussian processes (GP), multi-layer perceptron (MLP), long short-term memory (LSTM) networks, and graph neural networks (GNN), have been employed. Typically, as the algorithm progresses and a set of fully evaluated architectures becomes available, a meta-model is trained using architecture topology as features and validation accuracies as labels. This model is then used to predict the validation accuracy of yet-to-be-evaluated architectures. Notably, White et al. \cite{white2021powerful} demonstrated that augmenting the model with Jacobian covariance as an additional feature can enhance its performance by up to 20\%. Shen et al. \cite{shen2023proxybo} further extended this approach by integrating zero-cost proxies into Bayesian optimization, resulting in 3-5$\times$ speedups over previous methods.

Existing model-based predictor approaches exhibit notable biases, limiting their effectiveness to specific search spaces and requiring fully evaluated architectures as labels. The high initialization time is also a concern. Dudziak et al. \cite{dudziak2021brpnas} attempted to address this issue by leveraging the model's binary relation and training a prediction model through iterative data selection. However, their approach still requires over 6 GPU days to conduct the search process. Furthermore, due to simplistic architecture and feature embedding methods, such as one-hot encoding, these models cannot process previously unseen operators. These limitations prompt us to explore a more robust operator encoding method that enables a pre-trained prediction model to operate effectively in any architecture search space and accommodate unseen operators. Essentially, we investigate the viability of relying solely on pre-trained model-based prediction as a universal zero-cost proxy for guiding searches across diverse architecture spaces with negligible search time.

\begin{figure*}[t]
    \centering
        \includegraphics[width=\textwidth]{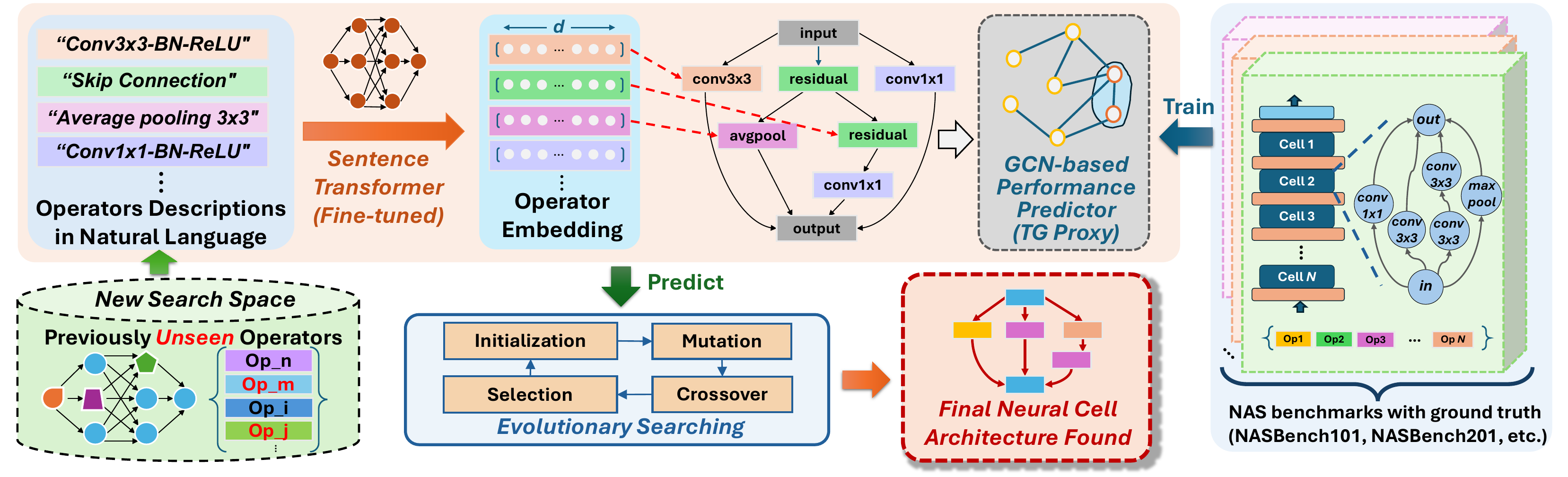}
    \caption{TG-NAS, featuring operator description embedding and GCN-based performance predictor}
    \label{fig:framework}
\end{figure*}

\section{TG-NAS: A Model-based Zero-cost Proxy}
In this work, we propose TG-NAS, a general model-based zero-cost proxy method for neural architecture search with a transformer embedding generator and a graph neural network predictor, supporting unseen operators in new search spaces. The embedding generator produces the desired feature embeddings with various techniques and the GCN predictor was trained with existing NAS benchmarks to predict the accuracy (or ranking) of a given architecture. Finally, the data-independent predicted ranking will guide the search algorithm as shown in Figure \ref{fig:framework}.

\subsection{Architecture Representation}
Various neural architecture search works represent their networks in the cell structure.
For instance, DARTS~\cite{liu2018darts}, and NATS-Bench (NAS-Bench-201)~\cite{dong2020nasbench201} define a cell-based search space representing each architecture as a directed acyclic graph (DAG), with nodes representing the features. NAS-Bench-101 \cite{ying2019nasbench101} on the other hand utilizes nodes to represent layers (operators) and edges for forward dataflow propagation. 
BANANAS \cite{white2020bananas} proposed a novel path-based encoding scheme and claimed it scales better than other methods.  Additionally, Yan et al. ~\cite{yan2021cate} propose a transformer-based encoding scheme with computation awareness. On contrary, our transformer-based neural architecture coding uses text-based DNN operator descriptions and sentence transformer. 

In this work, we represent the DNN architecture candidates using DAG with nodes representing operators and edges corresponding to model data propagation flow. We then represent the graphs with adjacent matrix and operator node embeddings, which becomes data input for our graph convolution network (GCN)\cite{kipf2017semisupervised} predictor. To make all search space comply with this representation, TG-NAS unifies other cell-based search space representations as shown in Figure \ref{fig:framework}. We applied this transformation to all NAS benchmark and search space in this work. NAS-Bench-101 was provided with the aforementioned architecture graphs, therefore no transformation is needed. Figure \ref{fig:framework} shows an example on NAS-Bench-201. The final model architectures of each search space are obtained by stacking multiple repeated cells with some other predefined cells in between. The differences between different DNN architecture candidates are purely determined by the cell architecture (represented as graph), so we use the embedding of individual cell architectures (as graphs) to represent the entire DNN architecture. Operator node features are encoded using a transformer model with a fixed length embedding size. The details can be seen in the following section.

\subsection{Operator Embedding Generator}

In our approach, we encode model cell graphs using an adjacency matrix together with node embeddings. However, encoding operators demand special attention as it involves representing and distinguishing various deep learning operators. Previous methods, which typically rely on one-hot vectors for operator encoding, are deemed suboptimal and non-portable, especially when dealing with unseen search spaces and operators.

Recognizing that the names of operators inherently contain valuable information, we assert that the \textbf{operator name} alone can provide insight into the operation. For instance, the operator name ``\textit{\textbf{CONV3x3-BN-ReLU}}'' suggests that it contains a two-dimensional convolution with a 3x3 kernel, followed by batch normalization and rectified linear activation. Therefore, we propose to construct a robust embedding model capable of extracting internal semantic information from operator names or their descriptive sentences in natural languages. For example, in the high-dimensional encoding space, operators like \textit{conv3x3} are expected to be closer to \textit{conv5x5} than to \textit{maxpool3x3}. Additionally, if the embedding model comprehends one type of operator, it should readily extend its knowledge to similar operators with, for example, different kernel sizes.
\renewcommand{\arraystretch}{1.1}
\begin{table*}[t]
\caption{Operator Descriptive Sentences Vary in Length for Generating Embedding}
\centering
\begin{adjustbox}{width=\textwidth}
\begin{tabular}{c|l|l|l}
\hline
\textbf{Operators} & \textbf{Short Sentences}        & \textbf{Medium-length Sentences}                             & \textbf{Long Sentences}                                                                                                                                                                                                    \\ \hline
\texttt{none}               & \textit{``None''}                & \textit{``Doing nothing''}                              & \textit{``A none operator that does nothing''}                                                                                                                                                                               \\
\texttt{skip\_connect}      & \textit{``Residual connection"} & \textit{``Identity mapping to the next layer"}         & \textit{``A residual connection operator that adds identity mapping to the next layer"}                                                                                                                                     \\
\texttt{nor\_conv\_3x3}     & \textit{``Convolution 3x3"}     & \textit{``Convolution 3 by 3 kernel, Batchnorm, ReLU"} & \textit{\begin{tabular}[c]{@{}c@{}}``A two-dimensional convolutional operator with a kernel size of 3 by 3 is applied, \\ succeeded by a batch normalization layer, and followed by a rectified linear layer"\end{tabular}} \\
\texttt{nor\_conv\_1x1}     & \textit{``Convolution 1x1"}     & \textit{``Convolution 1 by 1 kernel, Batchnorm, ReLU"} & \textit{\begin{tabular}[c]{@{}c@{}}``A two-dimensional convolutional operator with a kernel size of 1 by 1 is applied, \\ succeeded by a batch normalization layer, and followed by a rectified linear layer"\end{tabular}} \\
\texttt{avg\_pool\_3x3}     & \textit{``Average pooling 3x3"} & \textit{``Average pooling 3 by 3 kernel"}                  & \textit{``A average pooling operator with a kernel size 3 by 3"}                                                                                                                                                            \\ \hline
\end{tabular}
\end{adjustbox}

\label{tab:op_embeddings}
\end{table*}

\renewcommand{\arraystretch}{1.1}

\begin{table*}[t]
\caption{Different Combination of Sentences Transformer Model and Embedding Sentence Length}
\centering
\begin{adjustbox}{width=\textwidth}
\begin{tabular}{c|c|c|cccccc}
\hline
\multirow{2}{*}{Model} & \multirow{2}{*}{Model Size (MB)} & \multirow{2}{*}{Embedding Size} & \multicolumn{3}{c}{Kendall's $\tau$}                                                 & \multicolumn{3}{c}{Spearman's $\rho$}                          \\ \cline{4-9} 
                                           &                                  &                                 & \multicolumn{1}{c|}{Short} & \multicolumn{1}{c|}{Medium} & \multicolumn{1}{c|}{Long} & \multicolumn{1}{c|}{Short} & \multicolumn{1}{c|}{Medium} & Long \\ \hline
all-mpnet-base-v2                          & 420                              & 768                             & 0.48                       & 0.49                        & \multicolumn{1}{c|}{0.46} & 0.66                       & 0.67                        & 0.65 \\ \cline{1-3}
MiniLM-L6-v2                               & 80                               & 384                             & \textbf{0.56(0.60\textsuperscript{*})}              & 0.44                        & \multicolumn{1}{c|}{0.40} & \textbf{0.76(0.80\textsuperscript{*})}              & 0.62                        & 0.56 \\ \cline{1-3}
MiniLM-L6-v2-64                            & 81                               & 64                              & 0.36                       & 0.29                        & \multicolumn{1}{c|}{0.32} & 0.54                       & 0.43                        & 0.47 \\ \hline
\end{tabular}

\end{adjustbox}
\caption*{\footnotesize{*Fine Tuned with Augmented DNN Operator Descriptions}}
\label{tab:different_sentence}
\end{table*}

Certain existing works have attempted to construct embedding vectors from words or sentences, such as GloVe \cite{pennington2014glove}, or employed character embeddings to capture fine-grained semantic and syntactic regularities. However, our earlier experiments indicated that these methods face challenges when dealing with previously unseen words, particularly operators in our case. Consequently, we have opted for Sentence Transformer \cite{reimers2019sentencebert} as our primary method for generating desired operator embeddings. As illustrated in Figure \ref{fig:framework}, the Sentence Transformer utilizes siamese and triplet network structures to derive semantically meaningful sentence embeddings that can be compared using cosine similarity. A pooling operation is applied to the output of the pre-trained transformer model to obtain a fixed-size sentence embedding. We compute the mean of all output word vectors as the pooling strategy to generate the final operator embedding.

In our experiments, we explore three different sizes of pre-trained sentence transformer models and three distinct operator sentence lengths, crucial for embedding generation analysis. As outlined in Table \ref{tab:different_sentence}, we define and experiment with short, medium, and long operator descriptions as inputs to the embedding generator model. The test results on NAS-Bench-201 are presented in Table \ref{tab:different_sentence}. All three models undergo pretraining on the same 25 dataset collections \cite{reimers2019sentencebert}, containing over 1 billion training sentence pairs in total. The triplet objective function is employed during the pretraining phase. This function, given an anchor sentence $a$, a positive sentence $p$, and a negative sentence $n$, tunes the network to ensure that the distance between $a$ and $p$ is smaller than the distance between $a$ and $n$. Mathematically, the loss function can be expressed as:
\begin{equation}
\mathcal{L} = max(||s_a - s_p|| - ||s_a - s_n|| + \epsilon)
\end{equation} 
where $s_i$'s are the sentence embeddings for $a$, $n$, and $p$ respectively, and $||\cdot||$ denotes a distance metric and $\epsilon$ represents a margin. The chosen margin $\epsilon$ ensures that $s_p$ is at least $\epsilon$ closer to $s_a$ than $s_n$. In this context, the Euclidean distance serves as the distance metric, and $\epsilon$ is specifically set to 1 during training.

The \textit{all-MiniLM-L6-v2-64} entries in Table \ref{tab:different_sentence} are downsampled from the original \textit{all-MiniLM-L6-v2} model using principal component analysis (PCA) to achieve a 64-dimensional embedding vector length. The results of our GCN model prediction ranking (which we will discuss in the next section), trained with NAS-Bench-101 and tested on NAS-Bench-201, are presented in Table \ref{tab:different_sentence}. Notably, the combination of short sentence embedding and the pre-trained \textit{all-MiniLM-L6-v2} sentence transformer model, with the embedding length of 384, yields the best results for both Kendall's $\tau$ and Spearman's $\rho$ correlation coefficients. Our later experiments will adopt this combination.

Furthermore, to tailor the embedding space more closely to neural architecture search, we fine-tune the \textit{MiniLM-L6-v2} model on a task-specific similarity dataset, which is constructed from the PyTorch \texttt{torch.nn} documentation. Specifically, we parse operator classes and group them by functional categories (e.g., convolution, pooling, normalization, activation). For each operator class, we extract both the class name and its accompanying textual description. To enhance both the diversity and semantic richness of the dataset, we further employ GPT-4o to generate multiple augmented descriptions per class. These augmentations provide varied linguistic formulations that better reflect how operator functionality can be described in natural language, thus helping the model generalize across heterogeneous textual inputs during inference. We fine-tune the operator embedding model using supervised similarity pairs constructed based on functional categories. Pairs from the same operator class are labeled as similar, pairs from different classes within the same category as related, and pairs from distinct categories as unrelated. A cosine similarity loss is used to align the embedding space accordingly. Detailed definitions are provided in Appendix.

\begin{wrapfigure}{R}{0.5\textwidth}
    \begin{minipage}{0.5\textwidth}
\vspace{-2em}
\begin{algorithm}[H]
\small
\caption{TG-NAS Search Algorithm}
\label{alg:TG-NAS}
\begin{algorithmic}[1]
    \State \textbf{Input:} $NP$ population size
    \State $g \gets 0$
    \While{$|pop| < NP$}
        \State $pop_i \gets \text{random\_configuration}()$
        \State $pop'_i \gets \text{discretized\_architecture}(pop_i)$
        \State $fitness_i \gets \text{GCN\_Inference}(pop'_i)$
    \EndWhile
    \While{$g < g_{max}$}
        \State $V_g \gets \text{mutate}(pop_g)$
        \State $U_g \gets \text{crossover}(V_g, pop_g)$
        \State $U'_g \gets \text{discretized\_population}(U_g)$
        \State $fitness_g \gets \text{GCN\_Inference}(U'_g)$
        \State $pop_{g+1}, fitness_{g+1} \gets \text{select}(pop_g, U_g)$
    \EndWhile
    \State \Return $POP_g$
\end{algorithmic}
\end{algorithm}
\vspace{-2em}
\end{minipage}
\end{wrapfigure}

\subsection{GCN Proxy Predictor}
After completing the universal architecture encoding and operator feature embedding, we employ a three-layer graph convolution network (GCN) \cite{kipf2017semisupervised} as our prediction model. With the normalization trick, GCN can be defined as
\begin{equation}
H = X*_{G}G_{\Theta} = f(\bar{A}X\Theta)
\end{equation}

where $\bar{A} = \tilde{D}^{-\frac{1}{2}}\tilde{A}\tilde{D}^{-\frac{1}{2}}$, $\tilde{A} = A+ I_n$, and $\tilde{D_{ii}} = \sum_j\tilde{A}_{ij}$. To prevent overfitting to a particular training search space, we incorporate graph normalization and weight decay techniques. Our overarching objective is to deliver a universally applicable pre-trained predictor model that requires no tuning for new search spaces. Consequently, the GCN predictor model is subject to heavy regularization. An additional crucial factor influencing our choice of GCN over other prediction models is its capability to handle vast differences in architectures. Given the varying dimensions of the adjacency matrix (from unseen search spaces) and operator matrix (from unseen operators), GCN emerges as a suitable choice, demonstrating flexibility in accommodating diverse architectural structures. 


\subsection{Evolutionary Searching}


Evolutionary and genetic algorithms have been commonly used to optimize the NAS \cite{white2023neural}. To enhance the efficiency of the search, we adopt similar approach. We first \textbf{initialize} the entire population of architecture candidates with continuous parameters, where we map all operators evenly into the value between 0 and 1. Then \textbf{mutation} operation and \textbf{crossover} operation are performed to produce a new child. After generating all the offspring, the selection process will kick in using our proposed proxy to select the elite candidates for the next generation. In the end, we select the final returned architecture as the search result.

\begin{wrapfigure}{R}{0.6\textwidth}
    \centering
        \includegraphics[width=\linewidth]{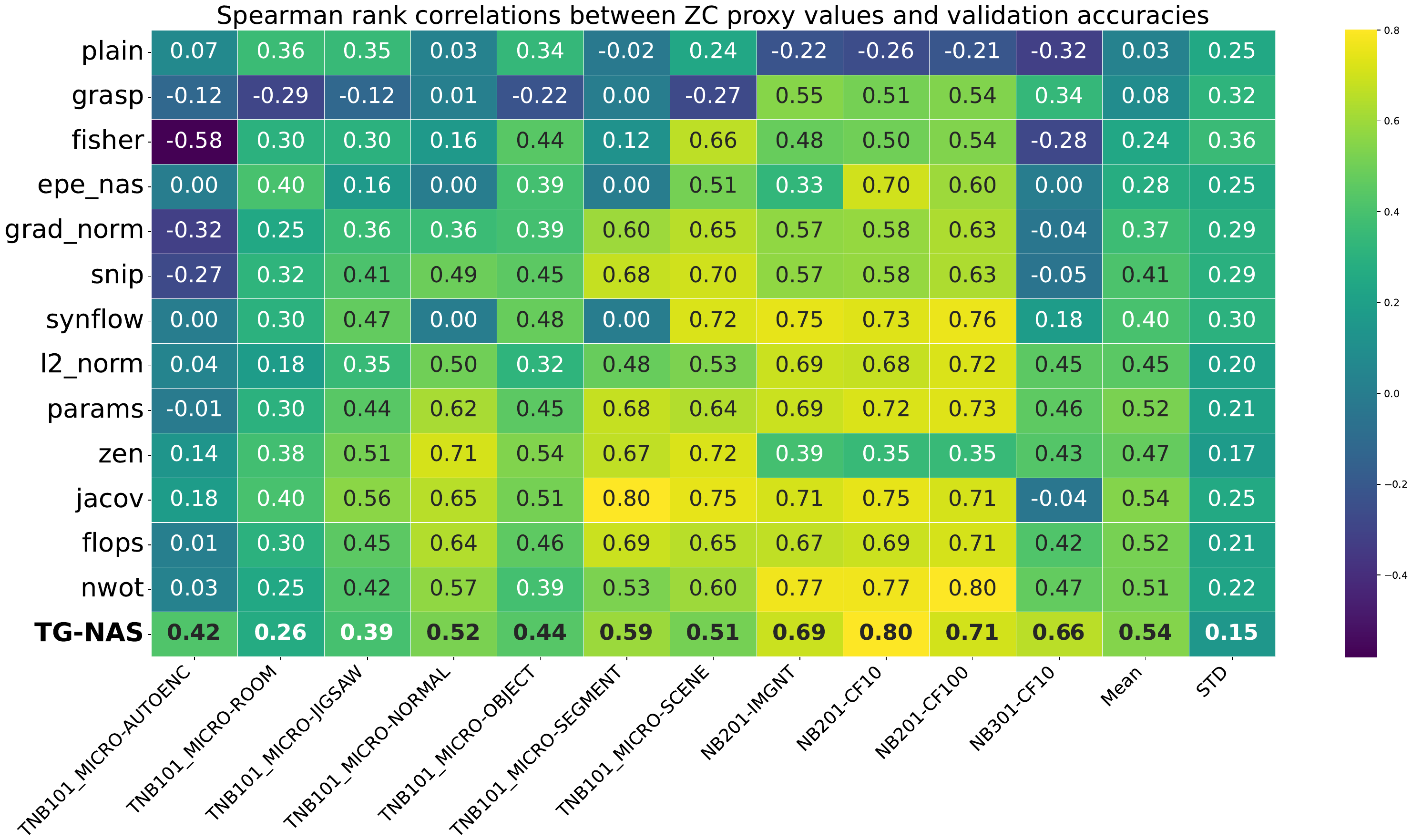}

    \caption{Spearman's $\rho$ rank between ZC proxy values and ground truth accuracies, for 14 ZC proxies and across 12 NAS benchmarks (the higher the better).}
    \vspace{-1em}
    \label{fig:zcp_corr_benchmarks}
\end{wrapfigure}

\section{Experiment and Results Analysis}
In our evaluation, we tested 12 NAS benchmarks similar to NAS-Bench-Suite-Zero \cite{krishnakumar2022nasbenchsuitezero}, including NAS-Bench-201 (CIFAR-10, CIFAR-100, and ImageNet16-120)~\cite{dong2020nasbench201}, NAS-Bench-301 (CIFAR-10)~\cite{zela2022surrogatenasbenchmarksgoing}, and TransNAS-Bench-101 Micro (Jigsaw, Object Classification, Scene Classification, Autoencoder, Room Layout, Surface Normal, and Semantic Segmentation)~\cite{duan2021transnasbench101improvingtransferabilitygeneralizability}. We excluded NAS-Bench-101~\cite{ying2019nasbench101} from the comparison because it was used as training data for our GCN predictor. To prevent information leakage and ensure a fair comparison, we removed it from subsequent evaluations.

For the GCN predictor, we employed a three-layer GCN with a weight decay of $3 \times 10^{-6}$ and an initial learning rate of 0.01, using the \textit{ReduceLROnPlateau} scheduler and the \textit{AdamW} optimizer. The model was trained for 150 epochs to ensure convergence. To minimize training cost and highlight the effectiveness of our novel architecture and operator encoding method in generalizing to unseen operators, the GCN was trained solely on NAS-Bench-101 (CIFAR-10)~\cite{ying2019nasbench101}.

\subsection{Generalizability Analysis of ZC Proxies Across 12 Benchmarks}

In Figure~\ref{fig:zcp_corr_benchmarks}, we report the Spearman's $\rho$ correlation between each zero-cost (ZC) proxy and the ground-truth performance for each benchmark. Among existing proxies, \textit{Jacov}\cite{mellor2021neural} and \textit{Zen Score}\cite{lin2021zennas} exhibit the highest rank correlations across all benchmarks. However, our proposed TG-proxy consistently outperforms them, achieving state of the art result.

Notably, in challenging tasks such as the TransNAS-Bench-101-Micro \cite{duan2021transnasbench101improvingtransferabilitygeneralizability} Autoencoder and Room Layout, where most ZC proxies perform poorly, TG-proxy maintains strong predictive power. On the widely used NAS-Bench-201\cite{dong2020nasbench201} benchmarks, although several proxies demonstrate good performance, TG-proxy still ranks the highest. While methods such as \textit{Snip}\cite{lee2019snip} and \textit{Grasp}\cite{wang2020picking} perform competitively on NAS-Bench-201, they generalize poorly to other benchmarks and are even outperformed by simple heuristics like \textit{Params} and \textit{FLOPs}.

Overall, these results highlight that TG-proxy not only achieves state-of-the-art performance on average accrossing search spaces but also with the lowest standard divination across diverse NAS benchmarks that demonstrates superior generalizability.

\subsection{Independent Analysis of proxies}
\begin{wrapfigure}{R}{0.6\textwidth}
    \centering
        \includegraphics[width=\linewidth]{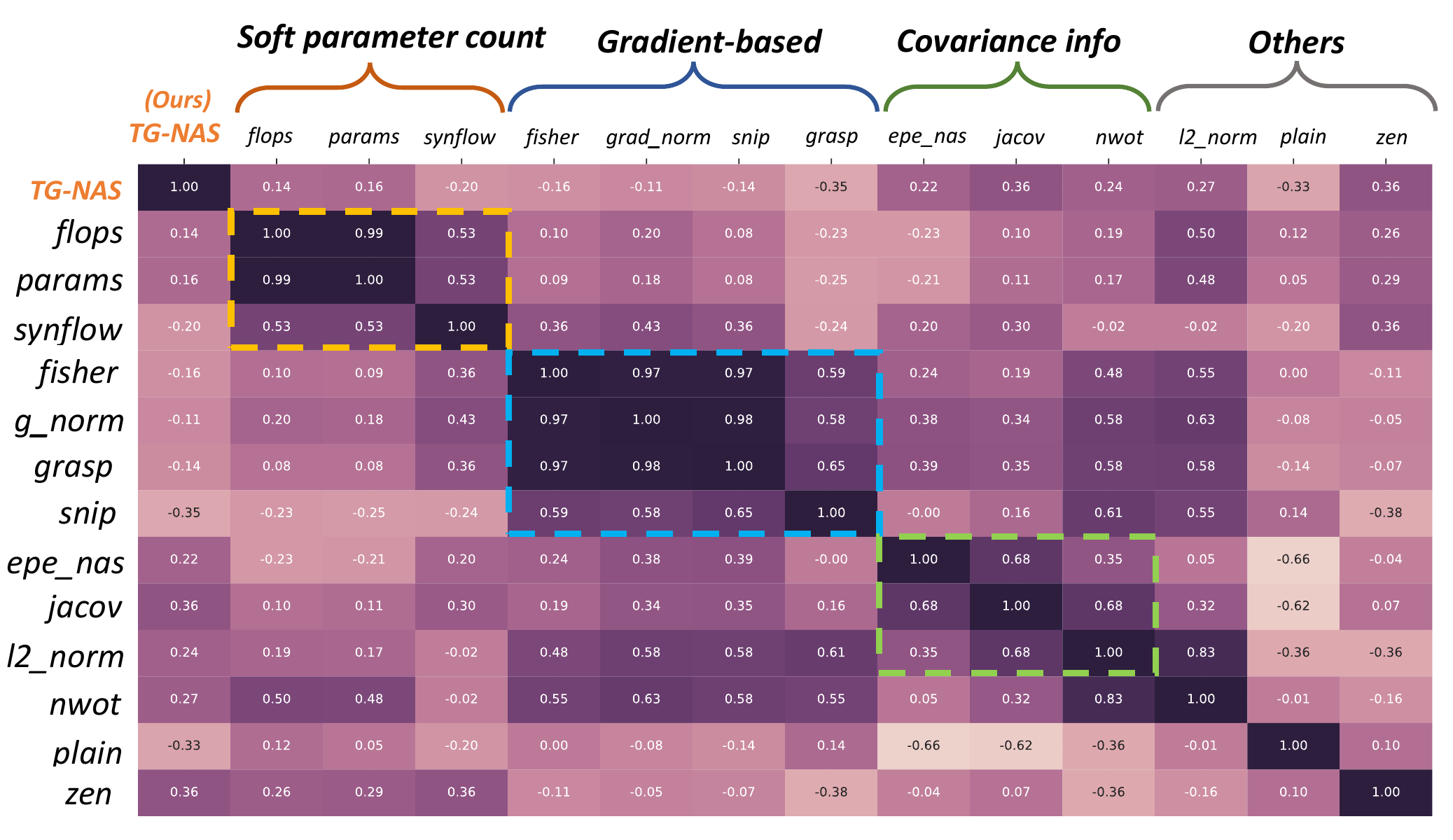}

    \caption{Spearman's $\rho$ Correlation between ZC proxy scores of CIFAR-10 on NAS-Bench-201.}
    \label{fig:zcp_corr}
\end{wrapfigure}


Multiple proxies could be combined to enhance the performance prediction accuracy. However, certain proxies could be highly correlated and do not provide extra information. Therefore, not all combinations provide good performance. We conducted assessments of zero-cost proxies on NAS-Bench-201 search spaces using the aforementioned three datasets. Additionally, by evaluating the same search space across different tasks, we aimed to determine if data- and task-independent zero-cost proxies offer universally applicable rankings. Our evaluation encompassed 14 zero-cost proxies, including our TG-NAS proxy result. We randomly sampled 1000 architectures from the search space and evaluated each zero-cost proxy metric for each architecture. Subsequently, we computed the \emph{Spearman's} rank correlation between each proxy.

Our analysis revealed a trend where the rank correlations of some zero-cost proxies were highly correlated, such as $FLOPs$ and $params$ , which is expected due to the relationship between parameters and computation in deep learning. This prompted us to compute the full correlations between all 14 pairs of zero-cost proxies and analyze their behaviors.

Figure \ref{fig:zcp_corr} shows the heatmap of correlations between pairs of popular proxies (calculated with CIFAR-10 results; CIFAR-100 and ImageNet16-120 results are available in the supplementary materials). We observed a consistent trend where $FLOPs$ and $params$ exhibited high correlations with each other. Additionally, $synflow$ \cite{abdelfattah2021zerocost} often showed high correlations with $FLOPs$ and $params$, consistent with recent work by Ning et al. \cite{ning2021evaluating}, which demonstrated that $synflow's$ value increases with the number of parameters in the architecture. We also noted high correlations between $grad\_norm$ \cite{abdelfattah2021zerocost}, $snip$ \cite{lee2019snip}, $grasp$ \cite{wang2020picking}, $nwot$ \cite{mellor2021neural}, and $fisher$ \cite{turner2020blockswap}, occasionally with $l2\_norm$ \cite{abdelfattah2021zerocost} as well, as they all leverage gradient saliency matrix information. Moreover, $epe\_nas$ \cite{Lopes_2021} and jacobian covariance ($jacov$) \cite{mellor2021neural} were highly correlated with each other.

It's intriguing to note that the $zen$ score \cite{lin2021zennas} and our TG-NAS proxy demonstrated the highest level of independence among all proxies evaluated. This suggests that our method offers a unique perspective on understanding model architectures. This discovery holds the potential to provide new insights into the design of zero-cost neural architecture search and the analysis of optimal combinations of zero-cost proxies. This could lead to further enhancements in zero-shot NAS techniques overall. We haven't included some zero-cost proxies due to various reasons: they either require training of supernet to make an evaluation or they did not release the source code \cite{dudziak2021brpnas, cai2019proxylessnas, shi2020bridging, xu2020pcdarts}.

\subsection{NAS Result on NAS-Bench-201}


\begin{figure}[htbp]
    \centering
    \begin{subfigure}[t]{0.30\textwidth}
        \centering
        \includegraphics[width=\textwidth]{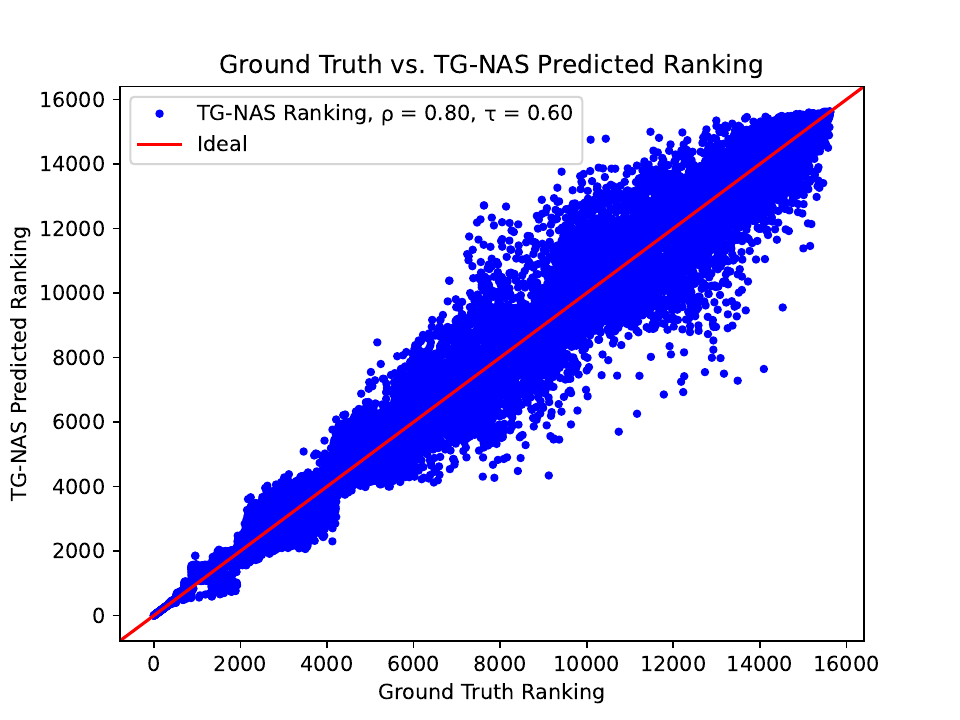}
        \caption{CIFAR-10}
    \end{subfigure}
    \hfill
    \begin{subfigure}[t]{0.30\textwidth}
        \centering
        \includegraphics[width=\textwidth]{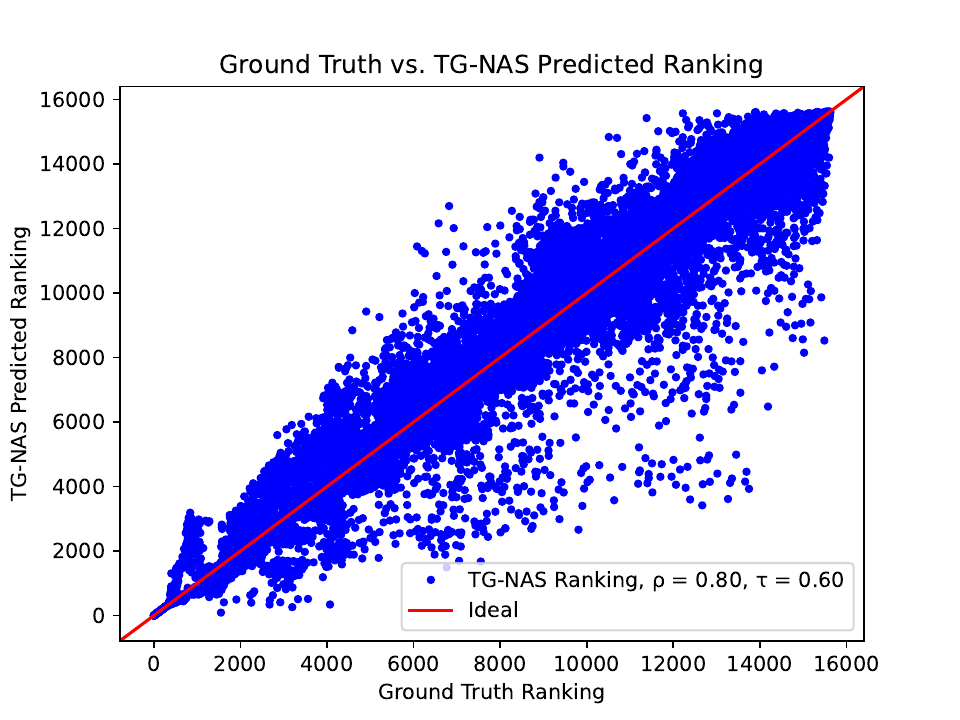}
        \caption{CIFAR-100}
        \label{reduction}
    \end{subfigure}
        \hfill
    \begin{subfigure}[t]{0.30\textwidth}
        \centering
        \includegraphics[width=\textwidth]{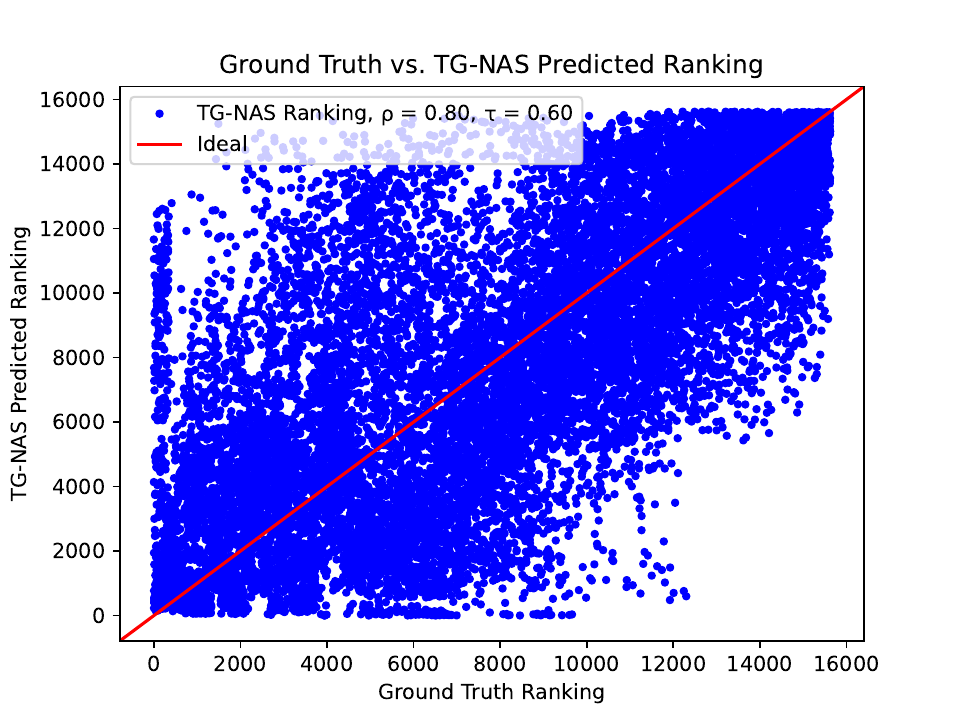}
            \caption{ImageNet16-120}
        \label{reduction}
    \end{subfigure}
    \hfill
    \caption{TG-NAS vs. Ground Truth ranking on NAS-Bench-201 Space}
    \label{fig:tg_prediction}
\end{figure}

Figure \ref{fig:tg_prediction} demonstrates that our TG proxy is highly positively correlated with the architecture’s accuracy ranking. Additionally, as illustrated in Table \ref{tab:table_cifar10}, our TG-NAS outperforms the majority of zero-shot NAS approaches and is comparable to the state-of-the-art TE-NAS results, achieving 93.75\% top-1 accuracy on CIFAR-10, while consuming only a fraction of the searching time of prior works.
\begin{table*}[t]
\caption{Results of CIFAR-10, CIFAR-100 and ImageNet16-120 on NAS-Bench-201}
\centering
\begin{adjustbox}{width=0.85\textwidth}
\begin{tabular}{lcccccccc}
\hline
  \multirow{2}{*}{Name of Works} &
  \multirow{2}{*}{\begin{tabular}[c]{@{}l@{}}FLOPs \\(M) \end{tabular}} &
  \multirow{2}{*}{\begin{tabular}[c]{@{}l@{}}Params \\(M) \end{tabular}} & 
  \multirow{2}{*}{\begin{tabular}[c]{@{}l@{}}Search Cost\\ (GPU Hours)\end{tabular}} &
  \multirow{2}{*}{\begin{tabular}[c]{@{}l@{}}CIFAR-10\\Accuracy (\%)\end{tabular}} &
  \multirow{2}{*}{\begin{tabular}[c]{@{}l@{}}CIFAR-100\\Accuracy (\%)\end{tabular}} &
  \multirow{2}{*}{\begin{tabular}[c]{@{}l@{}}ImageNet16-120\\Accuracy (\%)\end{tabular}} \\ 
 &               &              &             &           &               \\
\hline
$\mu$NAS\cite{muNAS}   & 7.78              & 0.073                      & 552            & 86.49          & 58.30   & 27.80           \\
DARTS\cite{liu2018darts}    & 82.49        & 0.587                  & 3.02           & 88.32          & 67.78         &34.60 \\
GDAS\cite{dong2019searching}   & 117.88     & 0.83                   & 8.03           & 93.36          & 69.64   &38.87      \\
KNAS\cite{knas}        & 153.27            & 1.073                 & 2.44           & 93.43          & 71.05    &45.05      \\
NASWOT\cite{mellor2021neural}  & 86.43     & 0.615                  & 0.09           & 92.96          & 69.70     &44.47     \\
TE-NAS\cite{chen2021neural}  & 188.66        & 1.317              & 0.43           & 93.78          & 70.44    &41.40     \\
\hline
\textbf{TG-NAS (ours)}    & \textbf{113.95} & \textbf{0.802}    & \textbf{0.01} & \textbf{93.75} & \textbf{70.64} & \textbf{44.97} \\ 
\hline
Ground Truth    & 153.27 & 1.073    & - & 94.37 &73.22 &46.71 \\ 
\hline
\end{tabular}
\end{adjustbox}


\label{tab:table_cifar10}
\end{table*}

\begin{table*}[!t]
\caption{Comparison with Recent NAS Works on ImageNet with the Mobile Setting}
\centering
\begin{adjustbox}{width=\textwidth}
\begin{tabular}{llllll}
\hline
\multirow{2}{*}{Name of Works}  & \multicolumn{2}{l}{Test Accuracy (\%)} & \multirow{2}{*}{\begin{tabular}[c]{@{}l@{}}Search Cost\\ (GPU Days)\end{tabular}} & \multirow{2}{*}{\begin{tabular}[c]{@{}l@{}}Params\\ (M)\end{tabular}} & \multirow{2}{*}{Search Method} \\ \cline{2-3}
                                & Top-1              & Top-5             &                                                                                   &                                                                       &                                \\ \hline
PNAS \cite{liu2018progressive}       & 74.2               & 91.9              & 225                                                                               & 5.1                                                                   & Bayesian Optimization                             \\
AmoebaNet-C \cite{real2019regularized} & 75.7               & 92.4              & 3150                                                                              & 6.4                                                                   & Evolution                      \\
NASNet-A \cite{zoph2018learning}   & 74.0               & 91.6              & 2000+                                                                             & 5.3                                                                   & Reinforcement Learning                             \\ \hline
DARTS \cite{liu2018darts}                           & 73.3               & 91.3              & 4.0                                                                               & 4.7                                                                   & Gradient-based                 \\
SNAS \cite{xie2018snas}                           & 72.7               & 91.8              & 1.5                                                                               & 4.3                                                                   & Gradient-based                 \\
BayesNAS \cite{zhou2019bayesnas}                        & 73.5               & 91.1              & 0.2                                                                               & 3.9                                                                   & Gradient-based                 \\
ProxylessNAS \cite{cai2019proxylessnas}                    & 75.1               & 92.5              & 8.3                                                                               & 7.1                                                                   & Gradient-based                 \\
TE-NAS \cite{dong2019searching}                          & 73.8               & 91.7              & 0.05                                                                              & 6.3                                                                   & Theoretical analysis           \\
BONAS-B \cite{shi2020bridging}                         & 75.2               & 92.7              & 5.0                                                                               & 4.5                                                                   & Bayesian Optimization                             \\ 
PC-DARTS \cite{xu2020pcdarts}                         & 74.9               & 92.2              & 0.1                                                                               & 5.3                                                                   & Gradient-based                             \\
PNASNet-5\cite{liu2018progressive}                          & 74.2               & 91.9              & 45                                                                               & 5.1                                                                   & Model-based Predictor                \\
GHN\cite{zhang2018graph} & 73.0               & 91.3              & 0.84                                                                              & 5.7                                                                   & Model-based Predictor                                     \\ 
NAONet\cite{luo2018neural}                        & 74.3               & 91.8              & 200                                                                               & 11.35                                                                   & Model-based Predictor                 \\\hline
 \textbf{TG-NAS (ours)}         & \textbf{74.9}      & \textbf{92.2}     & \textbf{0.0014}                                                                   & \textbf{5.6}                                                          & \textbf{Model-based Predictor} \\ \hline

\footnotesize{*All results are searched on CIFAR-10 and evaluated on ImageNet if not specialized noted}
\end{tabular}
\end{adjustbox}

\label{tab:ImageNetcomparision}
\end{table*}

Notably, despite the efficiency claims of zero-cost NAS methods over conventional NAS due to avoiding the training of sampled architectures, variations in computational cost and search time persist among them. For example, TE-NAS \cite{chen2021neural} required over 4 GPU hours, while ZiCo \cite{li2023zico} demanded over 10 GPU hours for the search on ImageNet on the DARTS space. Despite the GPU differences, this discrepancy arises from the fact that several proxy computations necessitate at least one forward pass or a forward-backward pass, requiring multiple runs for result stabilization. Additionally, many are data-dependent, limiting their general applicability and extending search times.

In contrast, TG-NAS completed the search in about 40 seconds, achieving up to a $300\times$ speedup compared to other zero-cost methods due to the lightweight nature of the proposed predictor model. Notably, the comparison of search times does not factor in the predictor model training time (the model training required 1.5 GPU hours), as it is a one-time training effort, and the pre-trained model can be distributed for use in various scenarios and search spaces.
\vspace{-1em}

\subsection{NAS Result for ImageNet on the DARTS Search Space}

\vspace{-1em}

For the DARTS search space, 
the final discovered cell architecture is shown in the Appendix. After the target cell was determined, the final network was constructed by stacking 14 cells, with the initial channel number set to 48. Performance results of TG-NAS discovered architectures compared against previous NAS works are shown in Table \ref{tab:ImageNetcomparision}. Our approach achieves a top-1/5 accuracy of 74.9\% and 92.2\% respectively. Notably, TG-NAS achieves up to 300$\times$ improvement in search efficiency compared to previous state-of-the-art zero-cost proxy methods. Furthermore, in comparison to other predictor-based NAS approaches, TG-NAS achieves over a 600$\times$ increase in search efficiency, completing the search process in less than two minutes on a single NVIDIA RTX 4090 GPU. 

\vspace{-1em}

\section{Conclusion}
\vspace{-1em}


In this work, we propose TG-NAS, a model-based zero-cost (ZC) proxy that is broadly applicable to new search spaces containing previously unseen operators. TG-NAS integrates text-based operator descriptions—processed by a fine-tuned sentence transformer—with a graph convolutional network (GCN) predictor, enabling it to function effectively as a ZC proxy. It offers key advantages in robustness, generalizability, proxy independence, and cost-effectiveness. Our experiments demonstrate that TG-NAS consistently outperforms existing proxies across a wide range of NAS benchmarks, establishing it as a strong foundational component for efficient architecture search. TG-NAS achieves up to 300× improvement in search efficiency over prior state-of-the-art ZC methods. Notably, it discover competitive models with 93.75\% CIFAR-10 accuracy on the NAS-Bench-201 space and 74.9\% ImageNet top-1 accuracy on the DARTS space.

\bibliographystyle{plain}
\bibliography{references}

\section{Appendix}
\subsection{GCN Proxy Predictor Functional Validation}

\begin{figure}[h!]
    \centering
    \includegraphics[width=\linewidth]{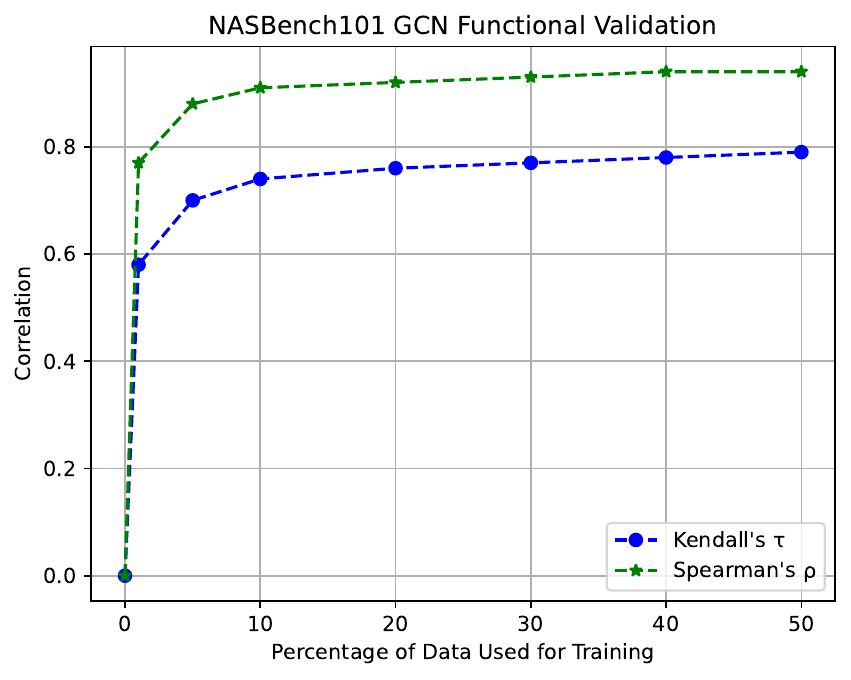}
    \caption{GCN Predictor Functional Validation on NAS-Bench-101 Benchmark}
    \label{fig:NASBench101_validation}
\end{figure}

To evaluate the applicability of the constructed GCN predictor, we partition the NAS-Bench-101 space into train/validation splits ranging from 90\% to 1\%. As shown in Figure~\ref{fig:NASBench101_validation}, the predictor maintains strong performance even when trained on as little as 1\% of the architecture data, as evidenced by high Kendall’s $\tau$ and Spearman’s $\rho$ correlation coefficients. Table~\ref{tab:zc_correlation} further compares these correlation scores with those of existing zero-cost proxies on NAS-Bench-201, highlighting the superior performance of our approach. Additionally, we investigate the impact of varying the number of GCN layers and training hyperparameters, with results summarized in Table~\ref{tab:gcn-kendall}.
\begin{table}[h]
\caption{\textit{Kendall's} $\tau$ and \textit{Spearman's} $\rho$ correlation between various zero-cost proxies on NAS-Bench-201}
\centering
\begin{adjustbox}{width=\textwidth}
\begin{tabular}{c|c|c|c|c|c|c|c|c}
\hline
Proxies & Params \cite{pham2018efficient} & FLOPs \cite{pham2018efficient}& SNIP \cite{lee2018snip} & Fisher \cite{turner2020blockswap} & Synflow \cite{tan2019mnasnet} & Zen-score \cite{lin2021zennas} & grad-norm  \cite{abdelfattah2021zerocost} & \textbf{TG-NAS} \\ \hline
\textit{Kendall’s} $\tau$         & 0.55   & 0.54  & 0.41 & 0.22   & 0.54    & 0.29      & 0.37      & \textbf{0.60}          \\ \hline
\textit{Spearman’s} $\rho$        & 0.74   & 0.73  & 0.58 & 0.36   & 0.73    & 0.38      & 0.54      & \textbf{0.80}          \\ \hline
\end{tabular}
\end{adjustbox}
\label{tab:zc_correlation}
\end{table}
\renewcommand{\arraystretch}{1.2}
\begin{table}[htbp]
\centering
\caption{Effect of Model Settings on Kendall’s $\tau$ Ranking}
\begin{tabular}{ccccc}
\toprule
\textbf{GCN Layers} & \textbf{Weight Decay} & \textbf{Sentence Length} & \textbf{Embedding Size} & \textbf{Kendall’s $\tau$} \\
\midrule
4 & $1\text{e}^{-4}$ & Long & 384 & 0.487 \\
4 & $1\text{e}^{-5}$ & Long & 384 & 0.401 \\
4 & $1\text{e}^{-6}$ & Long & 384 & 0.496 \\ \hline
4 & $1\text{e}^{-4}$ & Short & 384 & 0.495 \\
4 & $1\text{e}^{-5}$ & Short & 384 & 0.454 \\
4 & $1\text{e}^{-6}$ & Short & 384 & 0.497 \\\hline
4 & $1\text{e}^{-4}$ & Long & 768 & 0.433 \\
4 & $1\text{e}^{-5}$ & Long & 768 & 0.421 \\
4 & $1\text{e}^{-6}$ & Long & 768 & 0.454 \\\hline
4 & $1\text{e}^{-4}$ & Short & 768 & 0.433 \\
4 & $1\text{e}^{-5}$ & Short & 768 & 0.421 \\
4 & $1\text{e}^{-6}$ & Short & 768 & 0.454 \\\hline
3 & $1\text{e}^{-4}$ & Long & 384 & 0.483 \\
3 & $1\text{e}^{-5}$ & Long & 384 & 0.566 \\
3 & $1\text{e}^{-6}$ & Long & 384 & 0.577 \\\hline
\color{red}3 & \color{red}$1\text{e}^{-4}$ & \color{red}Short & \color{red}384 & \color{red}0.601 \\
3 & $1\text{e}^{-5}$ & Short & 384 & 0.599 \\
3 & $1\text{e}^{-6}$ & Short & 384 & 0.598 \\\hline
3 & $1\text{e}^{-4}$ & Long & 384 & 0.516 \\
3 & $1\text{e}^{-5}$ & Long & 384 & 0.557 \\
3 & $1\text{e}^{-6}$ & Long & 384 & 0.594 \\\hline
3 & $1\text{e}^{-4}$ & Short & 768 & 0.538 \\
3 & $1\text{e}^{-5}$ & Short & 768 & 0.545 \\
3 & $1\text{e}^{-6}$ & Short & 768 & 0.513 \\
\bottomrule
\end{tabular}
\label{tab:gcn-kendall}
\end{table}

\subsection{Additional Ranking Figures of TG-NAS vs. Ground Truth on NAS-bench-201}
In this section, we present additional accuracy and ranking vs. ground truth figures for CIFAR-10, CIFAR-100, and ImageNet16-120 datasets on the NAS-bench-201 space. The TG proxy GCN model utilized here is trained using the entire NAS-bench-101 benchmark with CIFAR-10 accuracy results only.
\begin{figure}[h!]
    \centering
    \begin{subfigure}[]{0.5\textwidth}
        \centering
        \includegraphics[width=\linewidth]{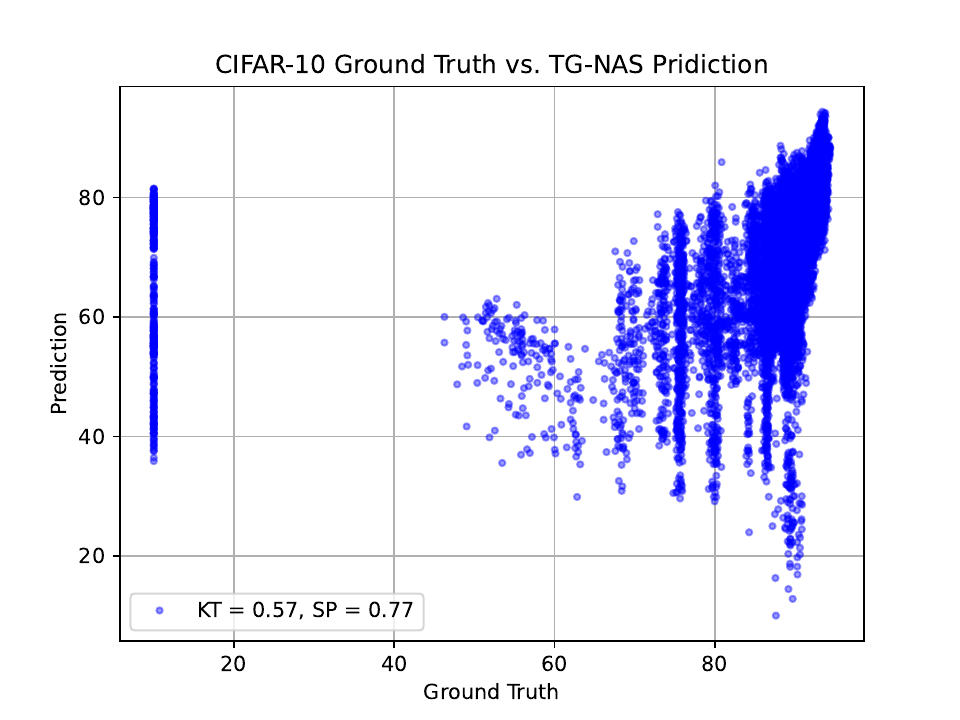}
        \caption{Ground Truth vs. TG-NAS Pridiction}
        \label{ntkvst}
    \end{subfigure}
    \hfill
    \begin{subfigure}[]{0.43\textwidth}
        \centering
        \includegraphics[width=\linewidth]{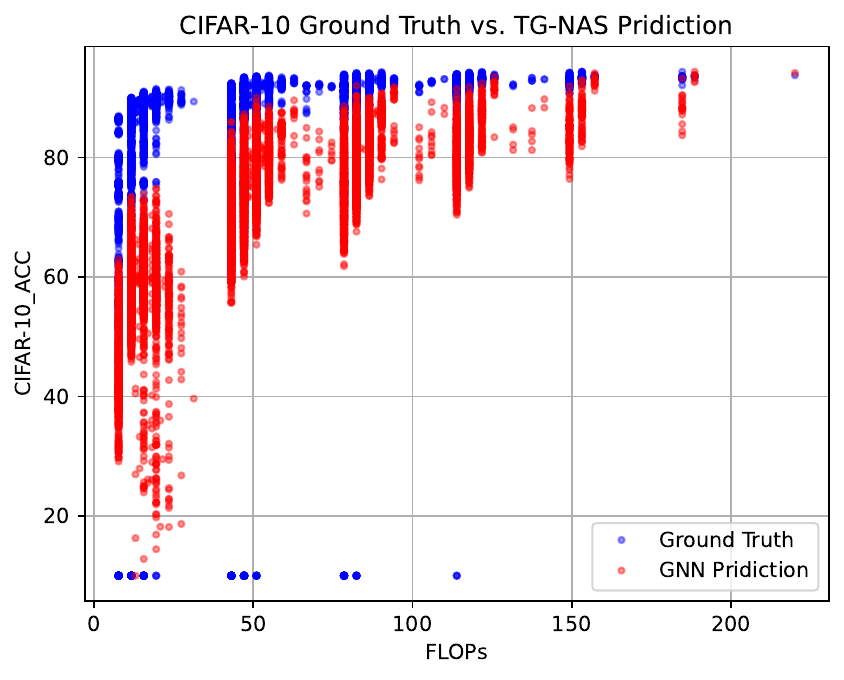}
        \caption{FLOPs vs. CIFAR-10 Accuracy}
        \label{batchvst}
    \end{subfigure}
    \caption{TG-NAS Accuracy Correlation Evaluation with Ground Truth on NAS-Bench-201 Space}
    \label{fig:tg_prediction}
\end{figure}

Figure \ref{fig:tg_prediction} demonstrates the correlation between our TG proxy predictions and the raw model accuracies, revealing a high positive correlation. This result is consistent with our previous ranking experiments.


We believe that incorporating more diverse benchmarks, including additional evaluation results from different datasets on the same architectures, will enhance the stability and generalizability of our TG-NAS approach.

\subsection{Additional Zero-cost Proxis Correlation Figures of CIFAR-100 and ImageNet16-120 Dataset on NAS-Bench-201}
In this section, we present heatmaps showing the correlations between pairs of popular proxies, calculated using the \textit{CIFAR-100} and \textit{ImageNet16-120} datasets as shown in figure\ref{fig:zcp_corr}. These heatmaps reveal correlation trends that are consistent with those observed in the \textit{CIFAR-10} results.
\begin{figure}[h!]
    \centering
    \begin{subfigure}[]{\textwidth}
        \centering
            \includegraphics[width=\linewidth]{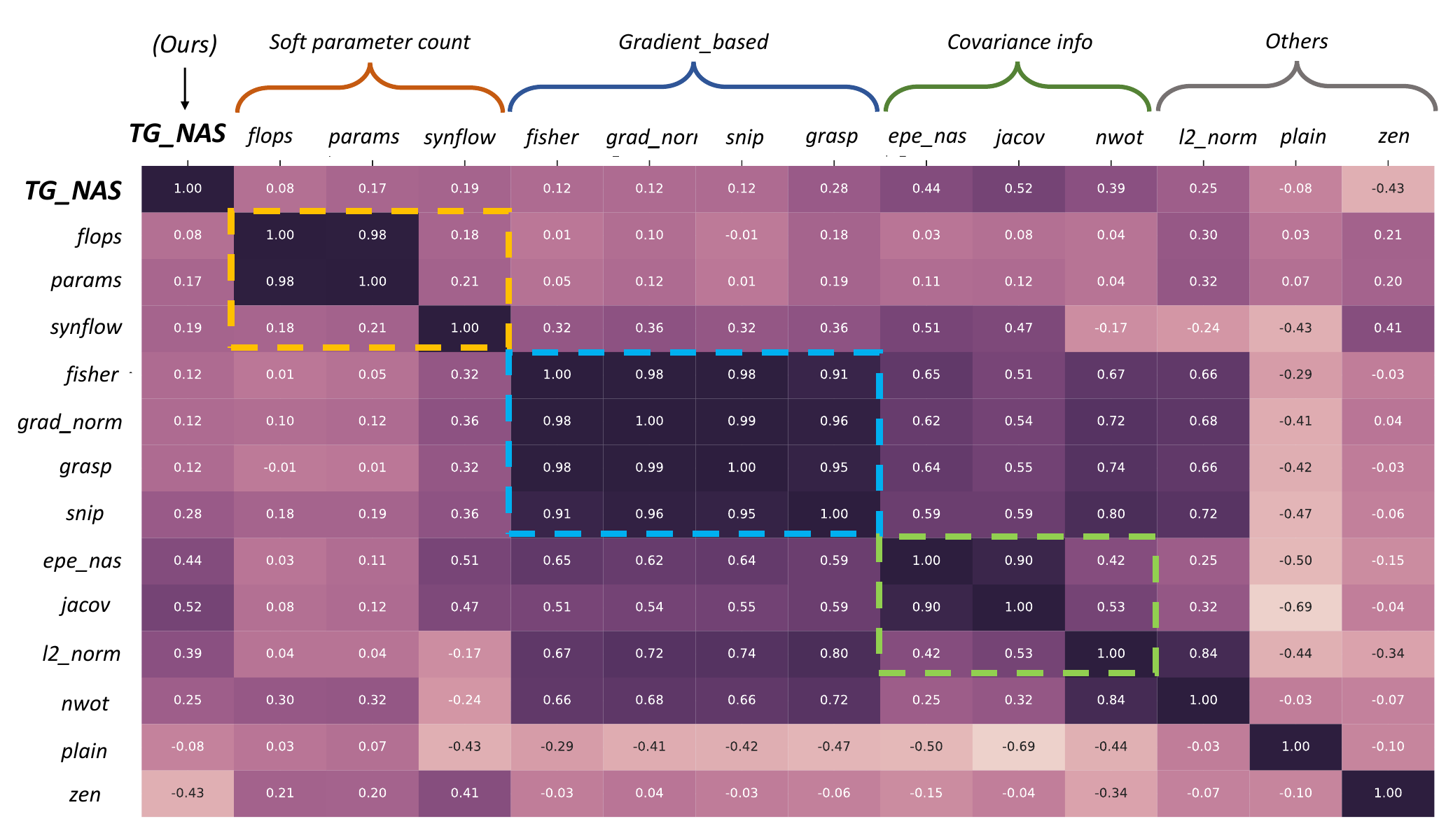}
        \caption{Spearman’s $\rho$ Correlation of CIFAR-100}
        \label{fig:zcp_corr_cifar-100}
    \end{subfigure}
    \hfill
    \begin{subfigure}[]{\textwidth}
        \centering
            \includegraphics[width=\linewidth]{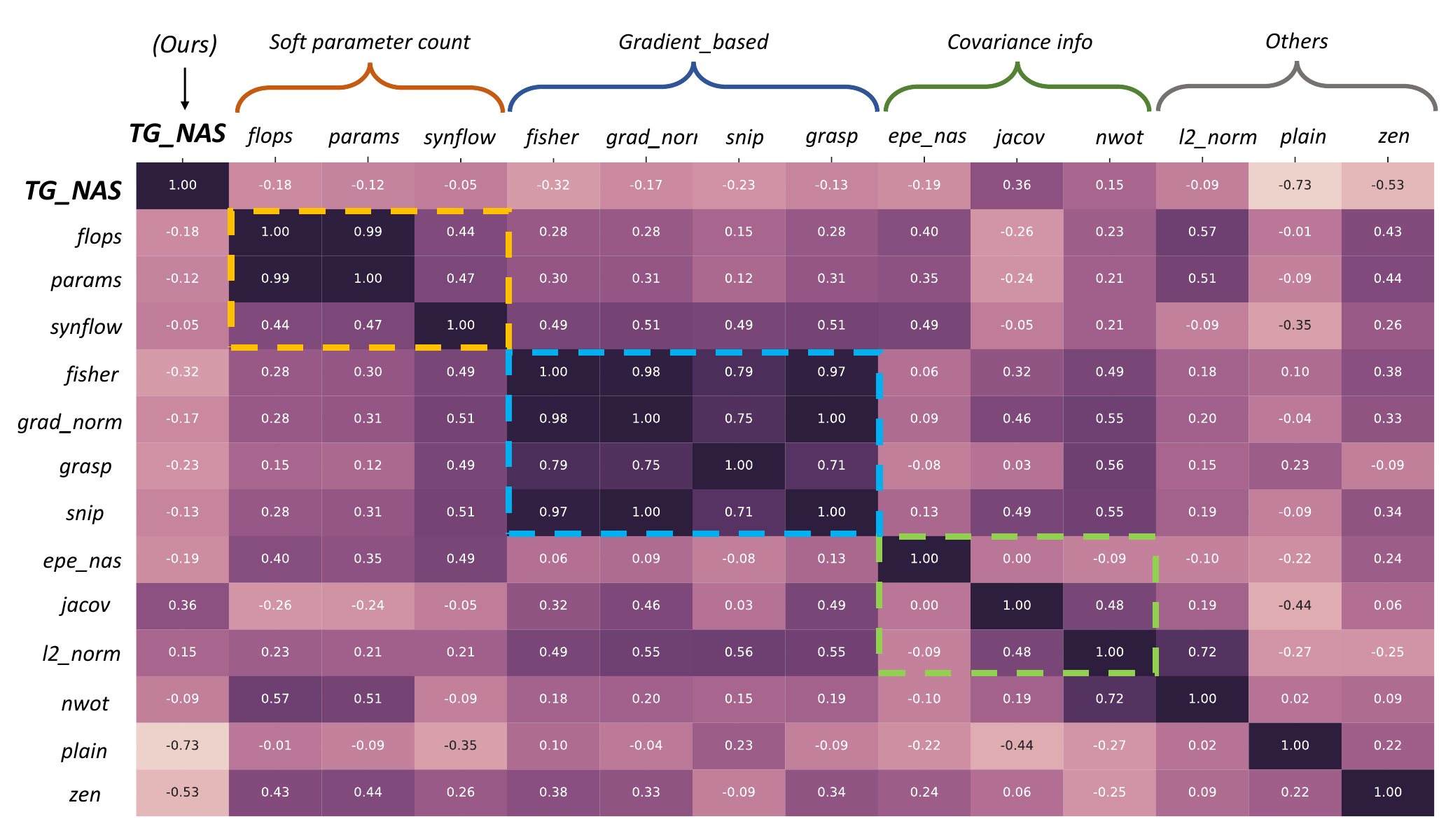}
        \caption{Spearman’s $\rho$ Correlation of ImageNet16-120}
        \label{fig:zcp_corr_imagenet16}
    \end{subfigure}
    \caption{Spearman’s $\rho$ Correlation for all Pairs of ZC Proxies of CIFAR-100 and ImageNet16-120 on NAS-Bench-201}
    \label{fig:zcp_corr}
\end{figure}

\subsection{Additional Comparison with More Predictor-based NAS Works}

In this section, we expand our comparison to include other NAS works involving training-based predictors. However, it's worth noting that the formulation and training of these predictor models differ from ours. They often require iterative training with new golden truth samples during the search process. Consequently, they are not zero-shot, highly bound to specific search spaces, lack general applicability, and remain costly to employ. To address this limitation, we propose the Transformer-based operator embedding method. This approach enables our predictor model to be decoupled from search spaces, thereby allowing our method to serve as a general zero-cost proxy. As illustrated in Table \ref{tab:ImageNetcomparision}, our TG-NAS achieves significantly higher search efficiency compared to other predictor-based NAS Works. Specifically, our method demonstrates search efficiency ranging from 71 times to $1.4 \times 10^5$ times better, highlighting its effectiveness and scalability in NAS tasks.

\begin{table*}[h!]
\caption{Comparison with additional predictor-based NAS works on ImageNet. The use of ``\textdagger'' indicates that the search is not conducted on the standard DARTS space, and we compare their results with similar mobile settings.}
\centering
\begin{adjustbox}{width=\textwidth}
\begin{tabular}{llllll}
\hline
\multirow{2}{*}{Name of Works}  & \multicolumn{2}{l}{Test Accuracy (\%)} & \multirow{2}{*}{\begin{tabular}[c]{@{}l@{}}Search Cost\\ (GPU Days)\end{tabular}} & \multirow{2}{*}{\begin{tabular}[c]{@{}l@{}}Params\\ (M)\end{tabular}} & \multirow{2}{*}{Search Method} \\ \cline{2-3}
                                & Top-1              & Top-5             &                                                                                   &                                                                       &                                \\ \hline
NGE\cite{li2020neural}     & 74.7               & 92.1              & 0.1                                                                               & 5.1                                                                   & Model-based Predictor                            \\
GHN\cite{zhang2018graph} & 73.0               & 91.3              & 0.84                                                                              & 5.7                                                                   & Model-based Predictor                                     \\ 
NAONet\cite{luo2018neural}                         & 74.3               & 91.8              & 200                                                                               & 11.35                                                                   & Model-based Predictor                 \\
PNASNet-5\cite{liu2018progressive}                          & 74.2               & 91.9              & 45                                                                               & 5.1                                                                   & Model-based Predictor                \\
GBDT-NAS-3S\textsuperscript{\textdagger} \cite{luo2020neural}                         & 76.5               & 93.2              & 4                                                                               & 6.4                                                                   & Model-based Predictor                 \\
CTNAS\textsuperscript{\textdagger} \cite{chen2021contrastive}                      & 77.3               & 93.4              & 50.1                                                                               & -                                                                   & Model-based Predictor                 \\
SemiNAS\textsuperscript{\textdagger} \cite{luo2020semi}                     & 76.5               & 93.2              & 4                                                                               &     6.3                                                               & Model-based Predictor                 \\
ZenNet-400M\textsuperscript{\textdagger} \cite{lin2021zennas}                         & 78.0               & -              &   0.5                                                                            & 5.7                                                                   & Zero Shot           \\
ZiCo-450M\textsuperscript{\textdagger}  \cite{li2023zico}                       & 78.1               & -              & 0.4                                                                               & 4.5                                                                   & Zero Shot                             \\ \hline
TG-NAS (ours)         & \textbf{74.5}      & \textbf{91.9}     & \textbf{0.0014}                                                                   & \textbf{5.6}                                                          & \textbf{Model-based Predictor} \\ \hline
\end{tabular}
\end{adjustbox}
\label{tab:ImageNetcomparision}
\end{table*}

\subsection{Sentence Transformer Finetune Setup}
To formulate supervised training pairs, we define three types of similarity relations:
\begin{itemize}
    \item Positive pairs (similarity = 1.0): Descriptions from the same class, including original and GPT-augmented versions (e.g., torch.nn.Conv2d, Pytorch offical description, and GPT4o augmented descriptions).
    \item Related pairs (similarity = 0.7): Descriptions from different classes within the same category (e.g., torch.nn.Conv2d and torch.nn.ConvTranspose2d).
    \item Unrelated pairs (similarity = 0.0): Descriptions from distinct functional categories (e.g., torch.nn.Conv2d vs. torch.nn.BatchNorm2d).
\end{itemize}
We fine-tune the model using cosine similarity loss, which encourages the embedding space to reflect semantic relationships: similar operators are embedded closer together, while unrelated ones are pushed apart. As shown in Table\ref{tab:sim-comparison}, this fine-tuning significantly improves the model’s ability to differentiate operator semantics. For example, the similarity score between conv2x2 and “A 2D conv layer with a 2x2 kernel” increases from 0.6052 to 0.8847, while the unrelated pair conv2x2 and maxpool drops from 0.2102 to 0.0275. This indicates the model learns to align functionally related operators while effectively distinguishing dissimilar ones.

\begin{table*}[h]
\caption{Operator Description's Similarity Comparison Before and After Finetuning }
\centering
\begin{adjustbox}{width=\textwidth}
\begin{tabular}{c|l|l|l}
\hline
\textbf{Operators} & \textbf{Compared Operators/Description} & \textbf{Similarity before Finetune} & \textbf{Similarity after Finetune} \\ \hline
conv2x2    & maxpool                   & 0.2102               & 0.0275               \\
conv2x2     & A 2D conv layer with a 2x2 kernel.  & 0.6052                 & 0.8847               \\ 
nn\_Dropout & nn\_BatchNorm2d              & 0.4629                  & -0.1160                  \\
skip connection     & residual                      & 0.0584              & 0.9113                   \\
maxpool     & avgpool                      & 0.2907              & 0.6741                  \\
maxpool     &conv2x2&0.2277&0.0011\\
\hline
\end{tabular}
\end{adjustbox}
\label{tab:sim-comparison}
\end{table*}

\begin{figure}[h]
    \centering
    \begin{subfigure}[t]{0.48\textwidth}
        \centering
        \includegraphics[width=\textwidth]{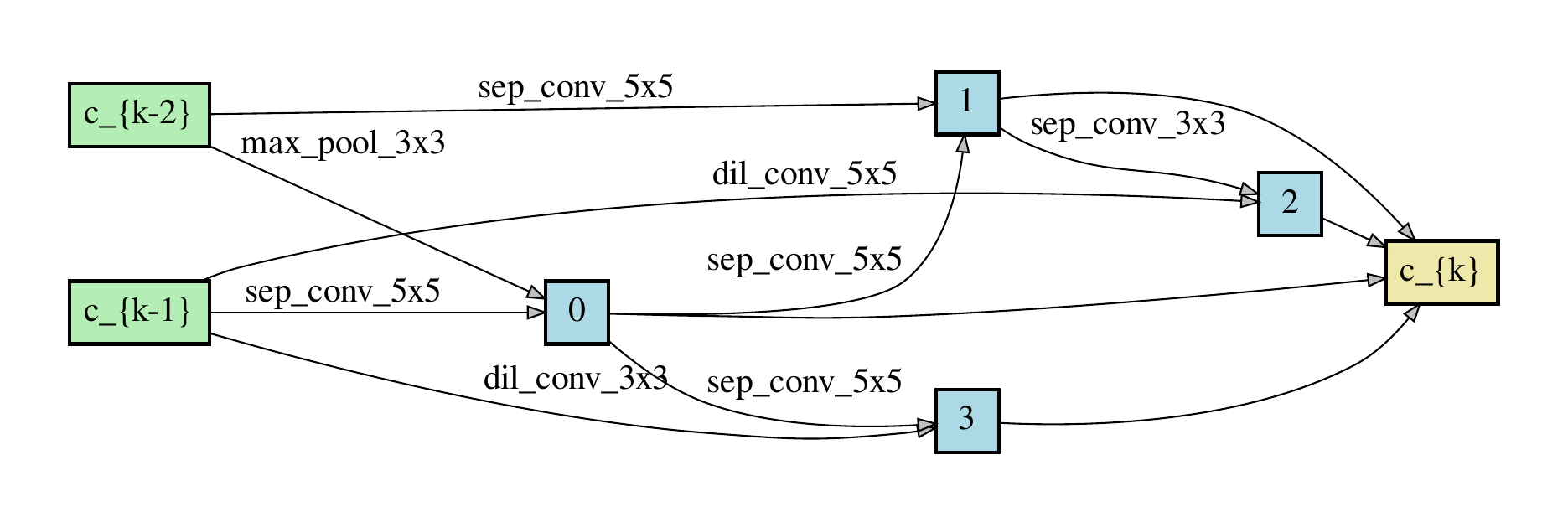}
        \caption{Normal Cell}
    \end{subfigure}
    \hfill
    \begin{subfigure}[t]{0.49\textwidth}
        \centering
        \includegraphics[width=\textwidth]{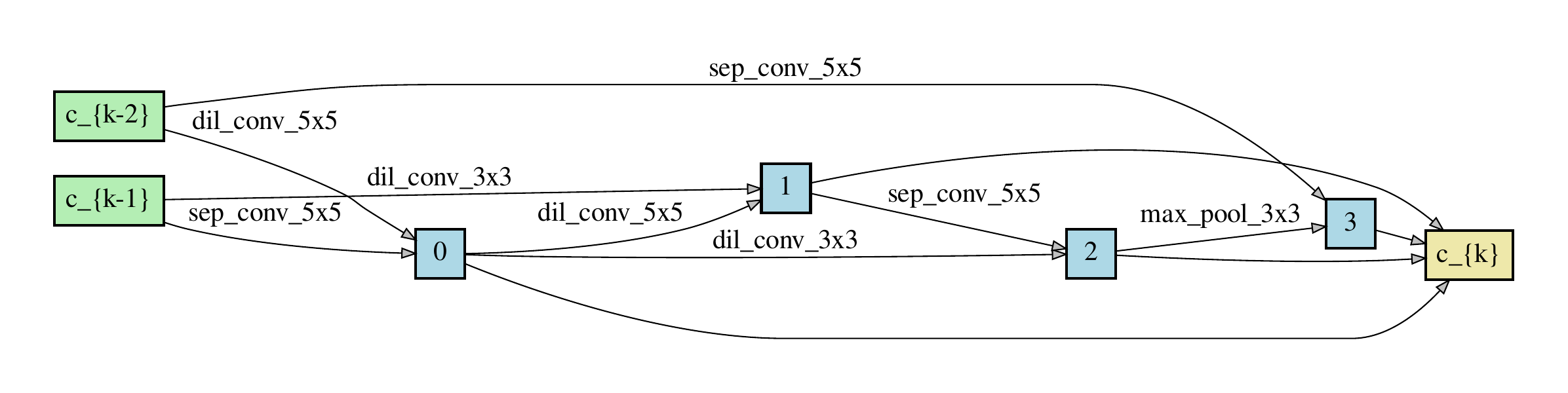}
        \caption{Reduction Cell}
        \label{reduction}
    \end{subfigure}
    \hfill
    \caption{Discovered Architecture on the DARTS Space}
    \label{fig:DARTS_result}
\end{figure}

\begin{figure}[htp]
  \centering
  \begin{subfigure}{0.4\textwidth}
    \centering
    \includegraphics[width=\linewidth]{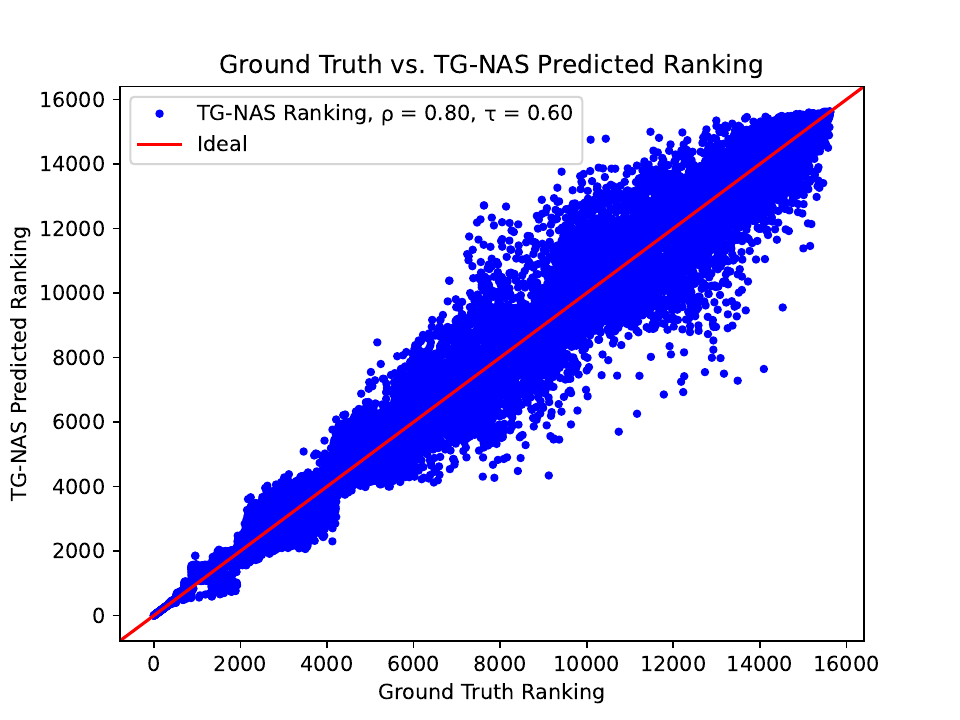}
    \caption{NB201-CF10}
  \end{subfigure}\hfill
  \begin{subfigure}{0.4\textwidth}
    \centering
    \includegraphics[width=\linewidth]{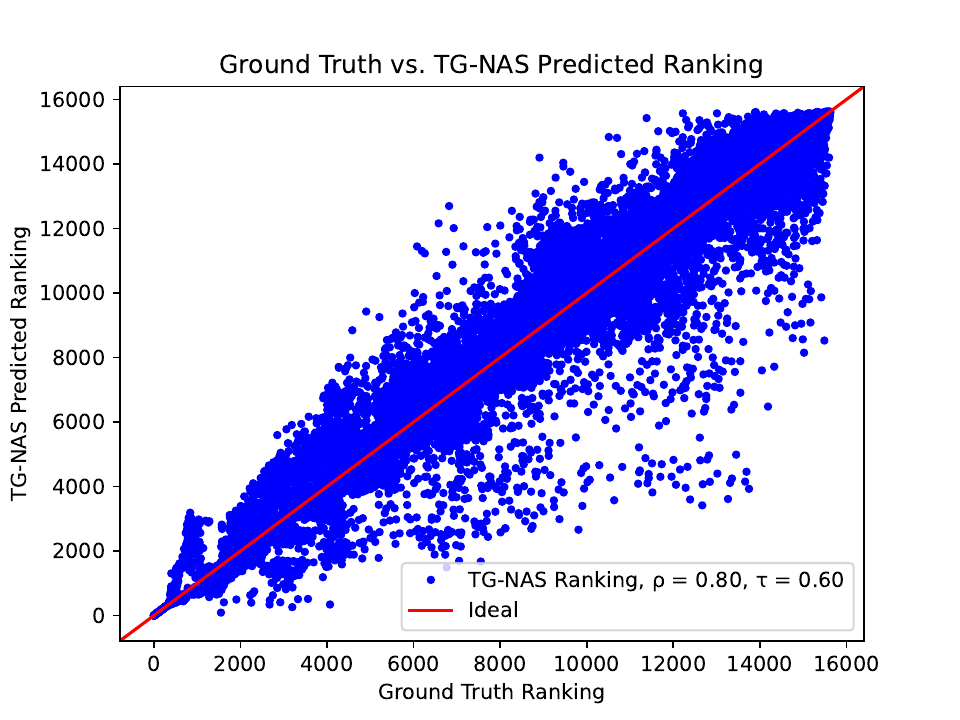}
    \caption{NB201-CF100}
  \end{subfigure}

  \vspace{1ex}
  \begin{subfigure}{0.4\textwidth}
    \centering
    \includegraphics[width=\linewidth]{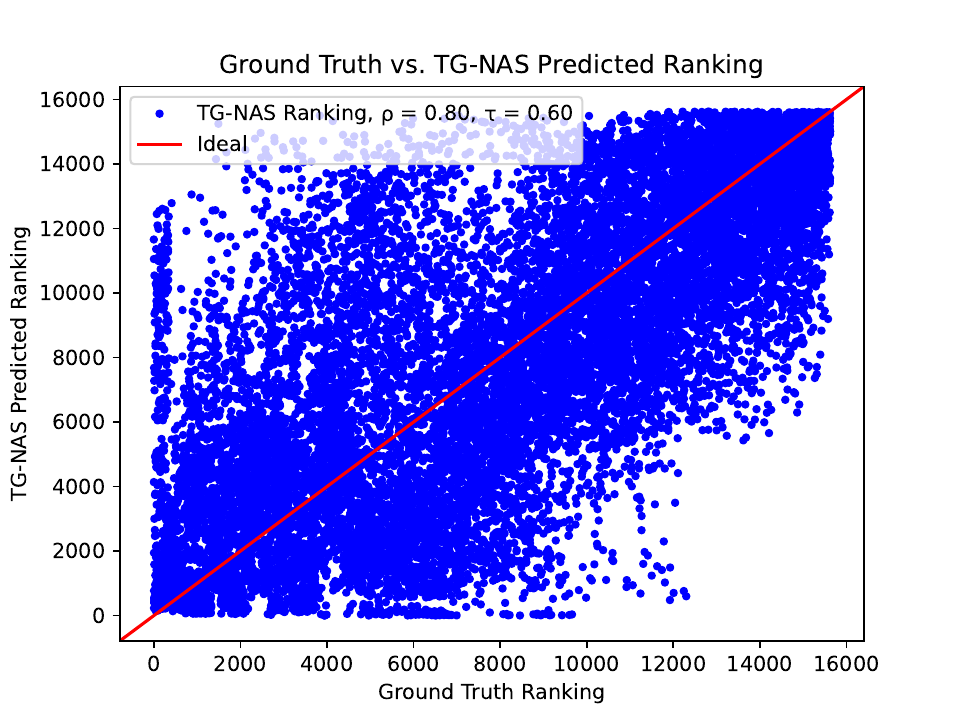}
    \caption{NB201-IMGNT}
  \end{subfigure}\hfill
  \begin{subfigure}{0.4\textwidth}
    \centering
    \includegraphics[width=\linewidth]{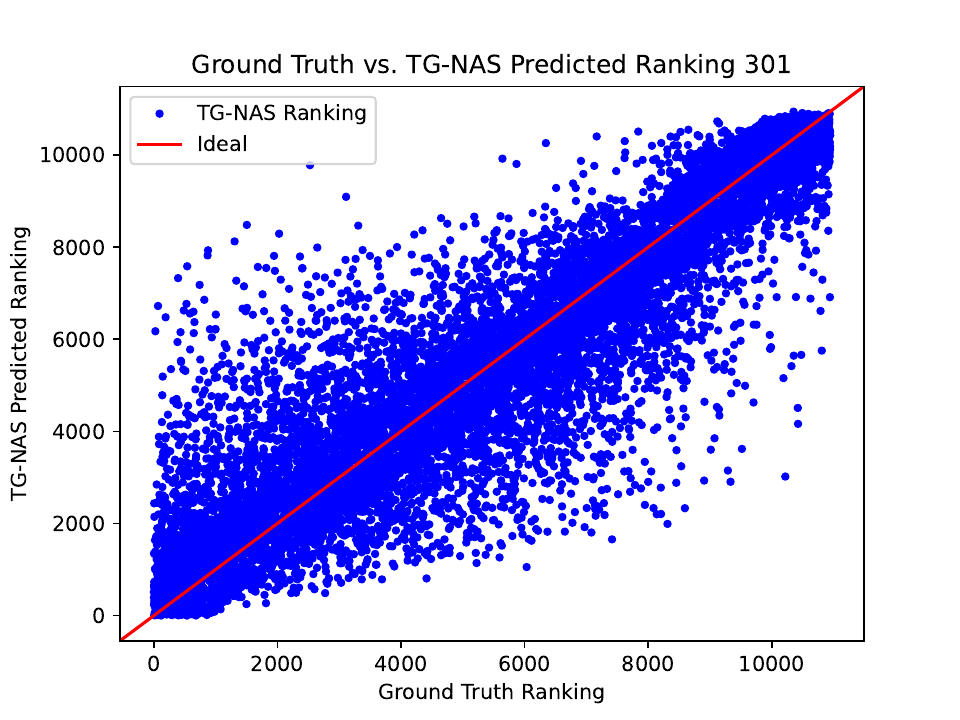}
    \caption{NB301-CF10}
  \end{subfigure}

  \vspace{1ex}
  \begin{subfigure}{0.4\textwidth}
    \centering
    \includegraphics[width=\linewidth]{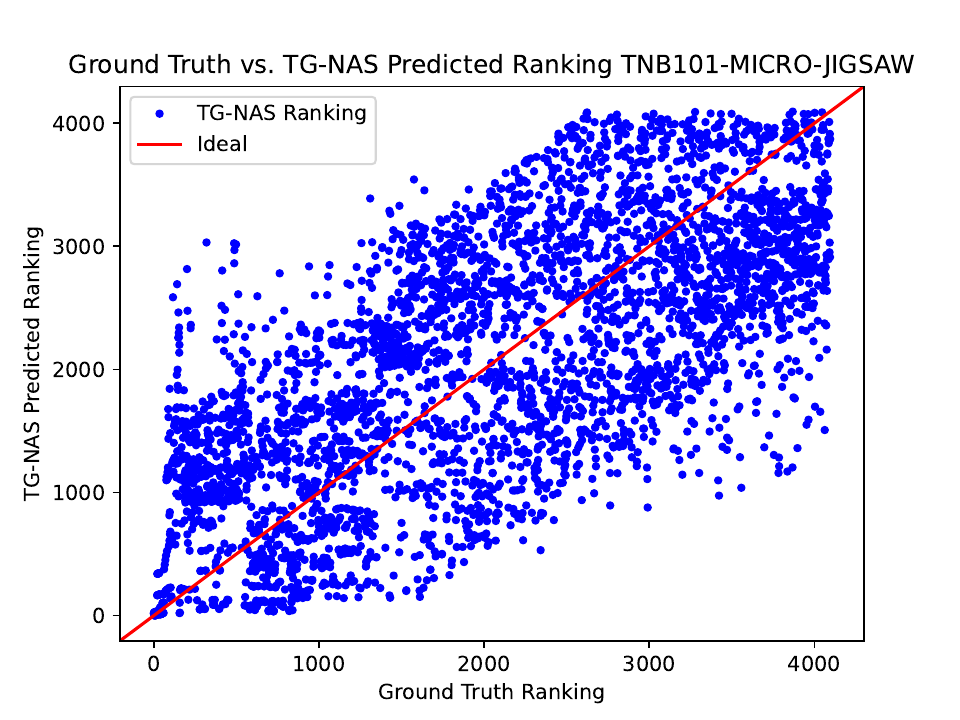}
    \caption{TNB101-MICRO-JIGSAW}
  \end{subfigure}\hfill
  \begin{subfigure}{0.4\textwidth}
    \centering
    \includegraphics[width=\linewidth]{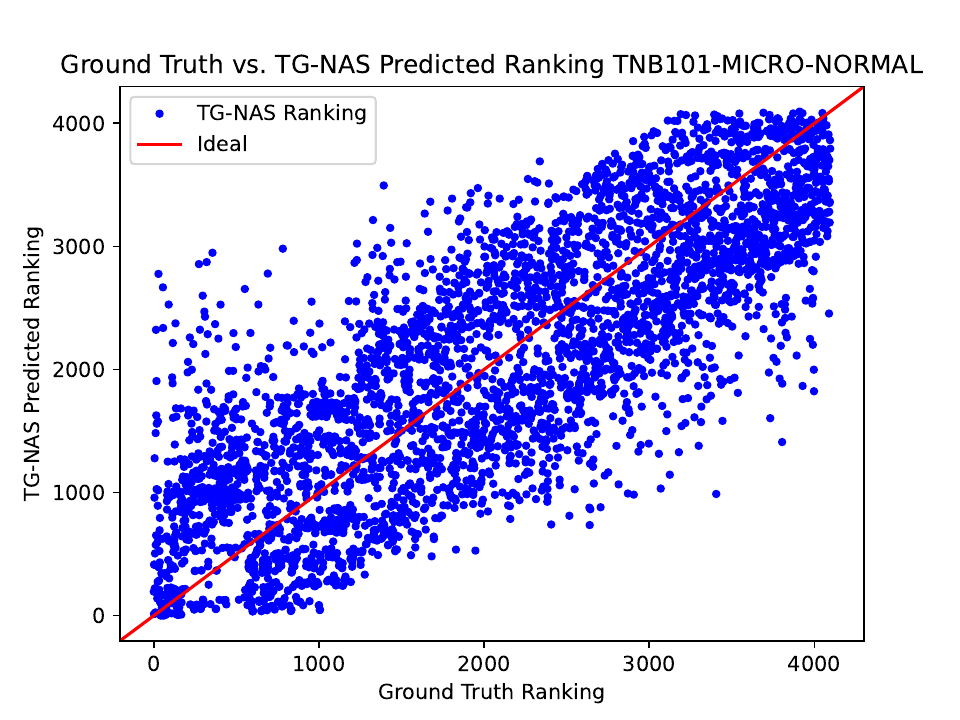}
    \caption{TNB101-MICRO-NORMAL}
  \end{subfigure}

  \vspace{1ex}
  \begin{subfigure}{0.4\textwidth}
    \centering
    \includegraphics[width=\linewidth]{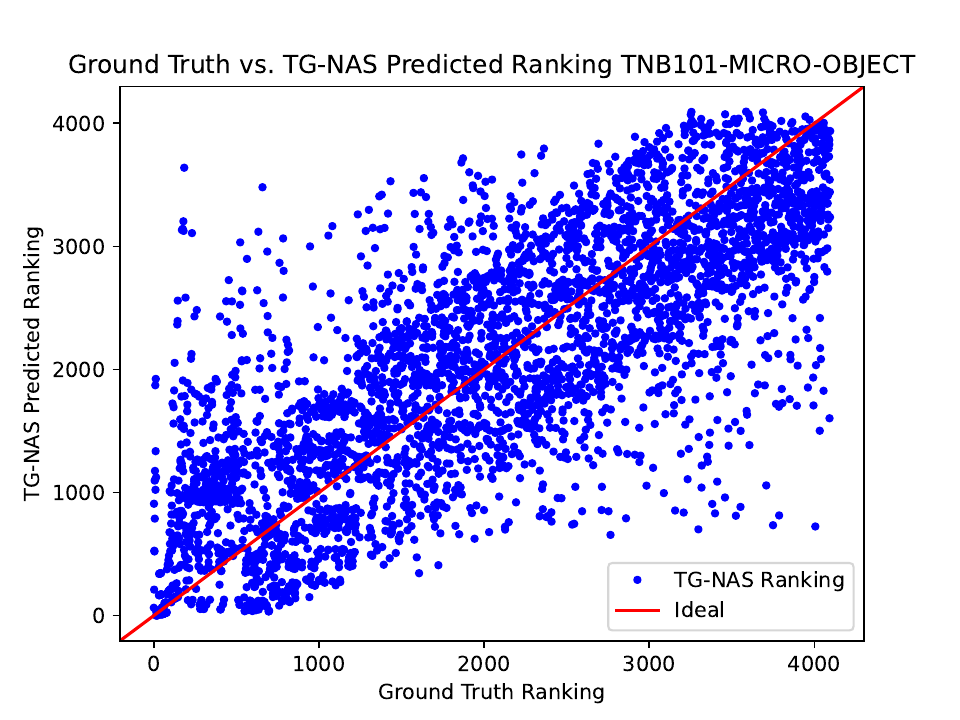}
    \caption{TNB101-MICRO-OBJECT}
  \end{subfigure}\hfill
  \begin{subfigure}{0.4\textwidth}
    \centering
    \includegraphics[width=\linewidth]{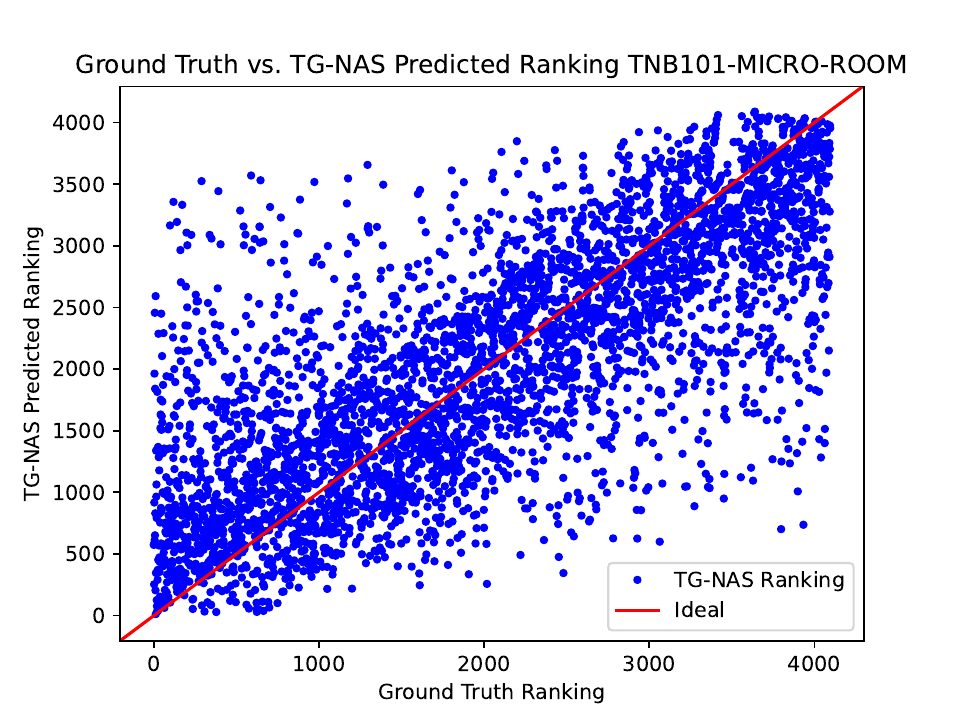}
    \caption{TNB101-MICRO-ROOM}
  \end{subfigure}

  \vspace{1ex}
  \begin{subfigure}{0.4\textwidth}
    \centering
    \includegraphics[width=\linewidth]{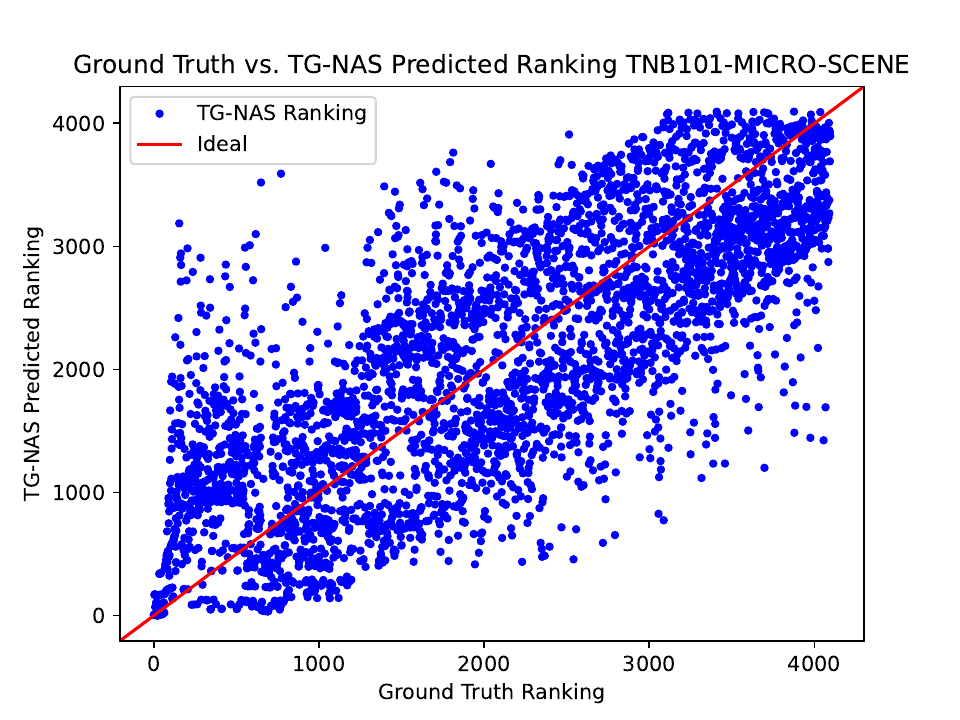}
    \caption{TNB101-MICRO-SCENE}
  \end{subfigure}\hfill
  \begin{subfigure}{0.4\textwidth}
    \centering
    \includegraphics[width=\linewidth]{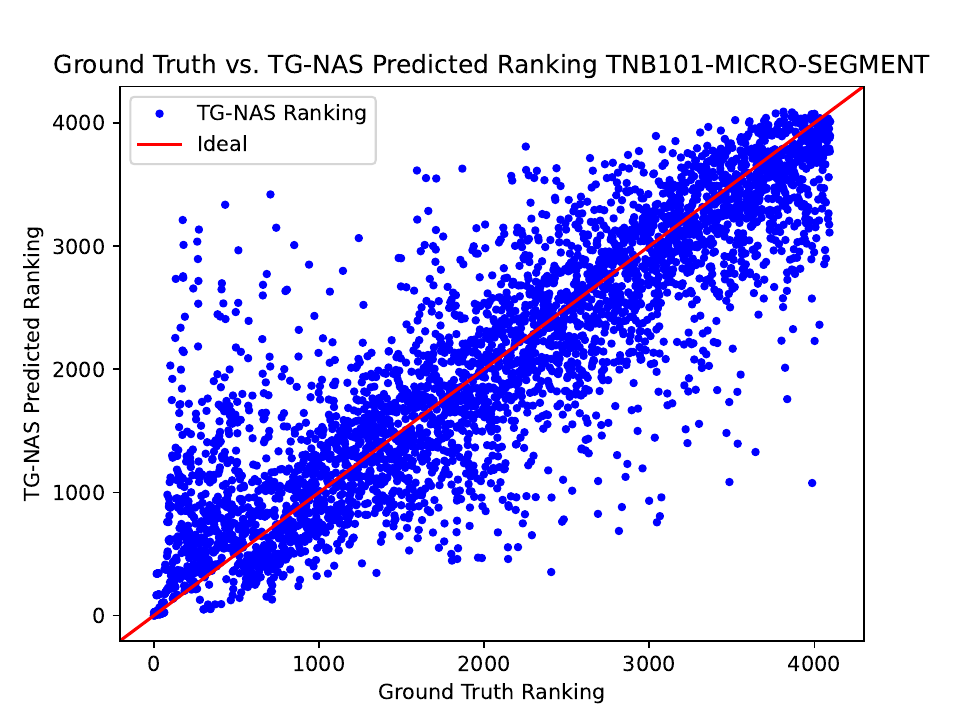}
    \caption{TNB101-MICRO-SEGMENT}
  \end{subfigure}

  \caption{Ground Truth vs Predicted Result on 10 datasets}
  \label{fig:comparison}
\end{figure}

\subsection{Final Searched Cell Architecture}
Our final search cell architecture on DARTS space can be found in figure\ref{fig:DARTS_result}

\subsection{Ground Truth and Predicted Result Comparison}
In Figure \ref{fig:comparison} we compare TG-NAS’s predicted ranking against the true ranking across ten benchmark tasks. The result on NAS-Bench-201 Cifar-10 and Cifar-100 exhibit very tight clustering, indicating good ranking fidelity. While the result for NAS-Bench-201 Imagenet shows relatively high variance, its Spearman's $\rho$ correlation is high as we have shown in the main paper. Overall, TG-NAS consistently captures the relative ordering of architectures across both large-scale image‐classification benchmarks and fine-grained micro-benchmarks.

\end{document}